\documentclass[10pt,twocolumn,letterpaper]{article}
\usepackage[T1]{fontenc} 

\usepackage{cvpr}              
\definecolor{cvprblue}{rgb}{0.21,0.49,0.74}
\usepackage[pagebackref,breaklinks,colorlinks,allcolors=cvprblue]{hyperref}
\usepackage{multirow}
\usepackage[table]{xcolor}         
\usepackage{graphicx}
\usepackage{amsmath}
\usepackage{bm}
\usepackage{caption}
\usepackage{subcaption}
\usepackage{arydshln}
\usepackage{tcolorbox}
\usepackage{tabularx}
\usepackage{tikz}
\usepackage{placeins}
\usepackage{stfloats}
\usepackage{cuted}
\usepackage[accsupp]{axessibility}  
\usetikzlibrary{arrows.meta}
\graphicspath{{./figs/}}

\def\paperID{46369} 
\def\confName{CVPR}
\def\confYear{2026}

\title{PaCo-RL: Advancing Reinforcement Learning for Consistent Image Generation \\ with Pairwise Reward Modeling}

\author{
Bowen Ping$^{1}$\footnotemark[1],\,
Chengyou Jia$^{1}$\footnotemark[1],\,
Minnan Luo$^{1}$\footnotemark[2],\,
Changliang Xia$^{1}$,\,
Xin Shen$^{1}$,\\
Zhuohang Dang$^{1}$,\,
Hangwei Qian$^{2}$\footnotemark[2]\\[3pt]
$^1$School of Computer Science and Technology, MOEKLINNS Lab, Xi'an Jiaotong University\\
$^2$CFAR and IHPC, A*STAR \\[3pt]
{\fontsize{8pt}{10pt}\selectfont  \texttt{jayceping6@gmail.com, cp3jia@stu.xjtu.edu.cn}}
}

\begin{document}
\twocolumn[{
\renewcommand\twocolumn[1][]{#1}
\maketitle
\begin{center}
    \vspace{-2em}
    \includegraphics[width=0.9\textwidth]{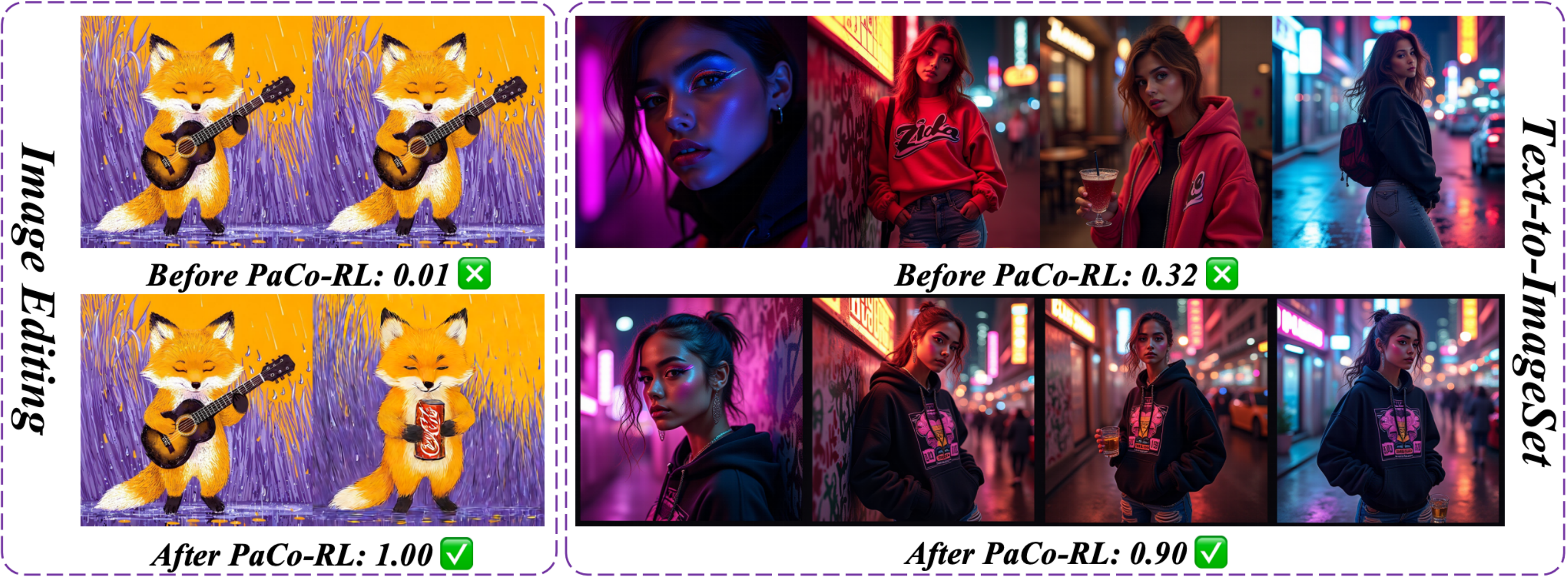}
    \vspace{-0.65em}
    \captionsetup{type=figure}
\caption{%
\textbf{Two Representative Tasks in Consistent Image Generation.} %
In the image editing task, %
the model needs to modify specific attributes while preserving the overall appearance. %
In the text-to-image set generation task, %
the goal is to generate multiple coherent images that remain consistent in identity, style, and context under a unified description.
}
\label{fig:teaser}
\end{center}
}]

\renewcommand{\thefootnote}{\fnsymbol{footnote}} 
\footnotetext[1]{Equal Contribution.}
\footnotetext[2]{Corresponding author.}

\begin{abstract}

Consistent image generation requires faithfully preserving identities, styles, and logical coherence across multiple images,
which is essential for applications such as storytelling and character design.
Supervised training approaches struggle with this task due to the lack of large-scale datasets capturing visual consistency and the complexity of modeling human perceptual preferences.
In this paper, we argue that reinforcement learning (RL) offers a promising alternative by enabling models to learn complex and subjective visual criteria in a data-free manner.
To achieve this, we introduce \textbf{PaCo-RL}, a comprehensive framework that combines a specialized consistency reward model with an efficient RL algorithm.
The first component, \textbf{PaCo-Reward}, is a pairwise consistency evaluator trained on a large-scale dataset constructed via automated sub-figure pairing.
It evaluates consistency through a generative, autoregressive scoring mechanism enhanced by task-aware instructions and CoT reasons.
The second component, \textbf{PaCo-GRPO},
leverages a novel resolution-decoupled optimization strategy to substantially reduce RL cost,
alongside a log-tamed multi-reward aggregation mechanism that ensures balanced and stable reward optimization.
Extensive experiments across the two representative subtasks show that PaCo-Reward significantly improves alignment with human perceptions of visual consistency,
and PaCo-GRPO achieves state-of-the-art consistency performance with improved training efficiency and stability.
Together, these results highlight the promise of PaCo-RL as a practical and scalable solution for consistent image generation.
\href{https://x-gengroup.github.io/HomePage_PaCo-RL/}{Project page}.

\end{abstract}    
\section{Introduction}
\label{sec:intro}

Image generation has made remarkable progress in producing high-quality and diverse images from textual prompts~\cite{saharia2022photorealistic,betker2023improving,esser2021taming,4o_image_generation,wu2025qwenimagetechnicalreport,jia2025chatgen,xia2025idealrealunifieddataefficient,jia2024ssmgspatialsemanticmapguided,wang2025jasmine,Wang_2024}.
However, achieving consistent image generation remains both challenging and crucial for various applications such as storytelling, advertising, and character creation~\cite{mao2024storyadaptertrainingfreeiterativeframework,tewel2024trainingfreeconsistenttexttoimagegeneration,chen2025interleavedscenegraphsinterleaved,zhou2024storydiffusion,liu2025onepromptonestory,huang2024incontextloradiffusiontransformers}.
In this work, we study consistent image generation through two representative tasks: Image Editing~\cite{luo2025editscoreunlockingonlinerl} and Text-to-ImageSet~\cite{jia2025settleonetexttoimagesetgeneration},
where the objective is to either align generated images with given references (\cref{fig:teaser}, left) or produce a coherent set of images from a single prompt (\cref{fig:teaser}, right).
This is a challenging problem for supervised training approaches,
due to the lack of large-scale datasets for visual consistency and the complexity of modeling human perceptions of consistency~\cite{wu2023humanpreferencescorev2,kirstain2023pickapicopendatasetuser,zhang2024learning,chen2025adiee,lin2024evaluating}.

Recently, reinforcement learning (RL) has demonstrated strong potential in advancing image generation models~\cite{liu2025flowgrpotrainingflowmatching,xue2025dancegrpounleashinggrpovisual,zheng2025diffusionnftonlinediffusionreinforcement}, offering a promising pathway to address the above challenges.
Instead of relying on explicit supervision from carefully curated datasets,
RL algorithms optimize generative models using feedback from reward models (RMs),
which are often trained to capture human preferences across diverse perceptual dimensions.
This enables models to learn complex and subjective visual criteria~\cite{xu2023imagereward,wu2023humanpreferencescorev2} in a data-free way.
Despite its promise, applying RL to consistent image generation remains an unsolved challenge, primarily due to the absence of both (1) \textbf{Suitable Reward Model} and (2) \textbf{Efficient RL Algorithm}.
On the reward side, existing reward models predominantly evaluate aesthetics and prompt alignment~\cite{ma2025hpsv3,wang2025unified,wu2025rewarddancerewardscalingvisual,wu2025editrewardhumanalignedrewardmodel},
but the lack of explicit consistency evaluation makes them insufficient for consistent generation.
On the optimization side, consistent generation is significantly more computationally demanding than single-image synthesis and requires carefully balancing consistency preservation and prompt fidelity.
However, current RL algorithms struggle to achieve this trade-off efficiently.

In this paper, we present a comprehensive methodology, \textbf{\underline{Pa}irwise \underline{Co}nsistency \underline{R}einforcement \underline{L}earning} (\textbf{PaCo-RL}), to overcome the above challenges.
It focuses on building a specialized, state-of-the-art consistency reward model and an efficient online RL algorithm for consistent image generation.
In summary, our contributions are:

(1) \textbf{PaCo-Reward.}
To build an effective and scalable reward model to assess visual consistency,
we first design an automated data synthesis pipeline based on sub-figure pairing,
which supports the construction of a large-scale, human-annotated consistency ranking dataset covering diverse image pairs with real-world consistency patterns.
Leveraging this dataset, we introduce PaCo-Reward,
which reformulates reward modeling as a generative task for pairwise comparisons.
Rather than relying on an extra regression head,
it aligns rewards with the next-token prediction process of the underlying vision-language model (VLM),
mapping consistency scores to the probability of generating the ``yes'' token.
To enhance generalization and interpretability,
PaCo-Reward incorporates task-aware instructions and CoT-style reasons,
providing a more robust and perceptually aligned consistency evaluator.

(2) \textbf{PaCo-GRPO.}
Recent RL methods for image generation~\cite{liu2025flowgrpotrainingflowmatching,xue2025dancegrpounleashinggrpovisual,li2025mixgrpounlockingflowbasedgrpo} achieve strong results but remain computationally heavy,
particularly for consistency tasks requiring high resolutions or multiple images per prompt.
To improve efficiency and stability, we introduce PaCo-GRPO, which integrates two core strategies:
(1) a resolution-decoupled training scheme that trains on low-resolution images, while preserving full-resolution generation at inference,
effectively reducing computational cost without compromising final performance;
and (2) a log-tamed multi-reward aggregation mechanism that balances rewards to prevent domination and ensure stable training.
Together, these techniques significantly boost training efficiency and stability without compromising performance,
and can be combined with prior optimizations~\cite{liu2025flowgrpotrainingflowmatching,li2025mixgrpounlockingflowbasedgrpo} to make consistent image generation both effective and computationally practical.

We conduct evaluations across multiple benchmarks for RMs,
where PaCo-Reward consistently outperforms existing reward models by a margin of 8.2\%-15.0\% in correlation with human preferences,
demonstrating its superior alignment with human perceptions of visual consistency.
Building on this strong reward foundation,
we integrate PaCo-Reward into our PaCo-GRPO,
which achieves state-of-the-art performance in both Text-to-ImageSet and Image Editing tasks,
yielding substantial improvements of 10.3\%-11.7\% in consistency metrics while preserving prompt fidelity,
together with nearly doubled training efficiency and greater stability by keeping the reward ratio under 1.8 to avoid domination.
These results underscore the broad potential of RL for advancing consistent image generation,
paving the way for future research in this important area.
\section{Related Work}
\label{sec:related_work}
\subsection{Consistent Image Generation}
\label{subsec:consistent_generation}
Consistent Image Generation aims to produce images coherent with given contexts, constraints, or attributes, encompassing tasks such as Text-to-ImageSet~\cite{tewel2024trainingfreeconsistenttexttoimagegeneration,hertz2024stylealignedimagegeneration,rahman2023make,jia2025settleonetexttoimagesetgeneration,song2025makeanythingharnessingdiffusiontransformers} and Image Editing~\cite{wang2025seededit30fasthighquality,wu2025editrewardhumanalignedrewardmodel,wu2025editrewardhumanalignedrewardmodel,luo2025editscoreunlockingonlinerl}
While differing in formulation, both share the goal of generating images faithful to shared identities, styles, or visual conditions.
Text-to-ImageSet methods rely on grouped or pairwise data with strong internal coherence~\cite{li2023photomaker,mao2024storyadaptertrainingfreeiterativeframework,liu2025onepromptonestory,huang2024consistentidportraitgenerationmultimodal,zhou2024storydiffusion,huang2024incontextloradiffusiontransformers},
whereas image editing approaches~\cite{batifol2025flux,wu2025qwenimagetechnicalreport,chen2025adiee,wang2026rl3dedit,wang2025editor} learn controlled modifications from paired original-edited images,
preserving fidelity while adjusting specific attributes.

Despite these differences, both task families face similar issues: dependence on curated paired or grouped data, narrow task-specific formulations, and limited generalization under complex or multi-condition constraints.
In this work, we unify these under a single \textit{Consistent Image Generation} framework, enabling PaCo-RL to enhance consistency across diverse generative scenarios.

\subsection{Reinforcement Learning for Image Generation}
\label{subsec:rl_image_generation}
\subsubsection{Reward Models for Image Generation}
\label{subsubsec:reward_models_image_generation}

Early works built reward models using existing image quality metrics or fine-tuned CLIP-based models on human preference datasets~\cite{wu2023human,wu2023human2,kirstain2023pickapicopendatasetuser,zhang2024learning,chen2025adiee,lin2024evaluating}.
Recent studies adopt multimodal large language models (MLLMs) as backbones for reward modeling~\cite{ma2025hpsv3,wang2025unified,wu2025rewarddancerewardscalingvisual,wu2025editrewardhumanalignedrewardmodel,luo2025editscoreunlockingonlinerl,seedream2025seedream40nextgenerationmultimodal},
demonstrating strong ability to understand complex human preferences and much greater flexibility across diverse evaluation tasks.
However, these works most remain focused on general image-text alignment or aesthetic evaluation, largely ignoring the aspects of visual consistency.

In contrast, consistency-related preferences require pairwise image comparisons rather than simple image-text alignment or single-image quality assessment.
Although models such as CLIP~\cite{radford2021learningtransferablevisualmodels} and DreamSim~\cite{fu2023dreamsim} can assess image similarity,
they are not designed to capture the multifaceted nature of human perception of consistency.
To bridge this gap, we propose a pairwise reward modeling framework that effectively captures human preferences for consistency across multiple aspects.


\subsubsection{RL Algorithms for Image Generation}
\label{subsubsec:rl_image_generation}
RL-style algorithms such as Proximal Policy Optimization (PPO)~\cite{schulman2017proximalpolicyoptimizationalgorithms,black2024training,fan2024reinforcement,gupta2025simple,miao2024training,zhao2025score},
Direct Preference Optimization (DPO)~\cite{rafailov2024direct,wallace2024diffusion,yang2024using,liang2024step,yuan2024self,liu2024videodpo,zhang2024onlinevpo,furuta2024improving,liang2025aesthetic},
and Group Relative Policy Optimization (GRPO)~\cite{shao2024deepseekmathpushinglimitsmathematical,liu2025flowgrpotrainingflowmatching,xue2025dancegrpounleashinggrpovisual,li2025branchgrpostableefficientgrpo,he2025tempflowgrpotimingmattersgrpo,li2025mixgrpounlockingflowbasedgrpo,wang2025coefficientspreservingsamplingreinforcementlearning} have emerged as powerful paradigms to enhance generative models~\cite{
peng2019advantage,fan2025online,wang2025prefgrpopairwisepreferencerewardbased,zheng2025diffusionnftonlinediffusionreinforcement,yeh2024training,tang2024tuning,song2023loss},
and have demonstrated significant success in aligning text-to-image models with human preferences~\cite{prabhudesai2023aligning,clark2023directly,xu2023imagereward,prabhudesai2024video,lee2023aligning,dong2023raft}.

Recent methods like Flow-GRPO~\cite{liu2025flowgrpotrainingflowmatching} and Dance-GRPO~\cite{xue2025dancegrpounleashinggrpovisual} introduce stochasticity to flow-matching models by converting ordinary differential equations (ODEs) into stochastic differential equations (SDEs), improving sampling diversity.
However, RL-based image generation remains computationally expensive, with \emph{sampling being a key bottleneck}.
Techniques such as MixGRPO~\cite{liu2025flowgrpotrainingflowmatching,li2025mixgrpounlockingflowbasedgrpo} reduce training costs by mixing SDE and ODE sampling,
but sampling overhead persists, particularly for consistent image generation tasks that handle multiple images.

To mitigate this, we propose PaCo-GRPO,
a resolution-decoupled training strategy that samples low-resolution images during RL, significantly reducing computational demands.
This approach complements existing methods like MixGRPO~\cite{li2025mixgrpounlockingflowbasedgrpo} and FlowGRPO-Fast~\cite{liu2025flowgrpotrainingflowmatching}, offering additional efficiency gains when combined.

\section{PaCo-Reward}
\label{sec:paco_reward}
To enable effective RL for consistent image generation,
we first build \textit{PaCo-Dataset}, a large-scale human-annotated dataset covering diverse consistency patterns,
designed for training and evaluating consistency reward models.
Based on this dataset, we introduce \textit{PaCo-Reward}, a novel framework for pairwise consistency reward modeling.
It reformulates consistency evaluation as a generative task,
accurately capturing human preferences for visual consistency while ensuring computational efficiency and interpretability.

\subsection{Paco-Dataset: Pairwise Consistency Dataset}
\label{subsec:consistencyrank}
\begin{figure*}[!t]
    \centering
    \begin{minipage}{0.85\textwidth}
        \centering
        \includegraphics[width=0.9\linewidth]{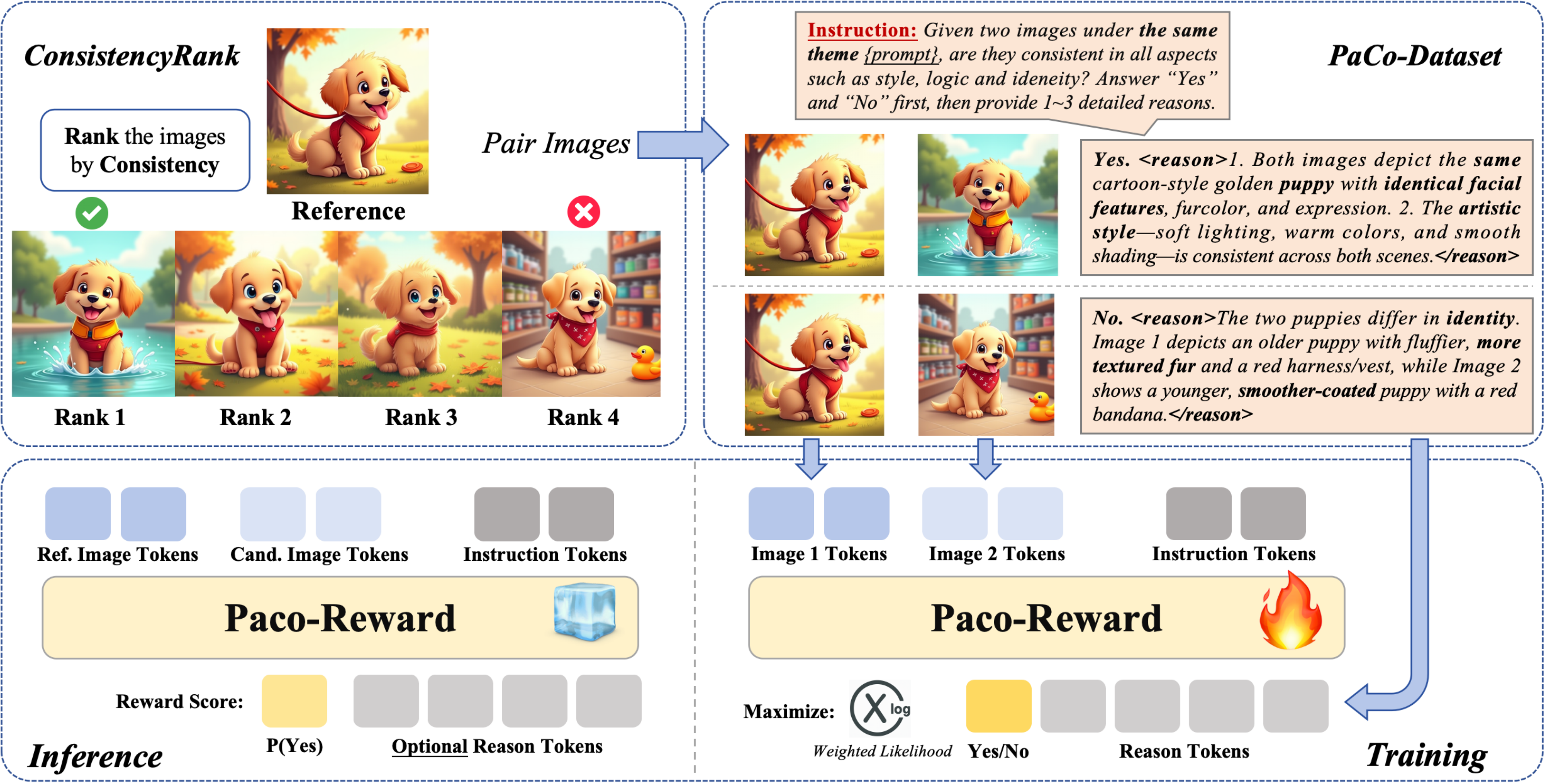}
        \vspace{-0.7em}
        \caption{
        Overview of the proposed \textbf{PaCo-Reward} framework.
        }
        \label{fig:pipeline}
    \end{minipage}
    \vspace{-1.5em}
\end{figure*}

\textbf{Overview.}
To address the shortage of high-quality datasets for consistency reward modeling,
we introduce Paco-Dataset, which spans 6 major categories and 32 subcategories of consistent image generation scenarios,
including character generation, design style generation, and process generation,
covering diverse types of visual consistency such as identity, style, and logic.
The full categorization is provided in Appendix~\ref{app:paco_dataset_details}.
Each entry consists of one reference image and four comparison images,
accompanied by human-annotated rankings reflecting visual consistency.
This dataset enables more effective training of reward models aligned with nuanced human preferences and includes a benchmark,
\textit{ConsistencyRank}, for standardized evaluation.


\noindent%
\textbf{Data Synthesis.}
Consistent generation involves multiple images sharing common elements while varying in other aspects,
making data collection particularly challenging.
To efficiently construct a large-scale dataset with diverse consistency patterns,
we first generate 2,000 text prompts for Text-to-ImageSet synthesis using Deepseek-V3.1~\cite{deepseekai2025deepseekv3technicalreport},
then select 708 diverse prompts through graph-based diversification on text embeddings~\cite{radford2021learningtransferablevisualmodels}.
Inspired by~\cite{huang2024incontextloradiffusiontransformers},
we use FLUX.1-dev~\cite{blackforestlabs_flux1dev_2024} to generate $m\times n$ image grids with strong internal consistency.
To enhance diversity and reduce cost, we generate four grids with different seeds per prompt
and apply a \emph{sub-figure combinatorial pairing} strategy, where we divide each grid into $m\times n$ subfigures and exhaustively pair them across grids from the same prompt.
We empirically choose a $2\times 2$ grid setting as it offers the best trade-off between quality and efficiency.
This process yields 33,984 unique ranking instances from 708 prompts and 2,832 images,
substantially expanding dataset scale and diversity at minimal cost.
Each instance comprises one reference image and four comparison candidates to be ranked by consistency,
as illustrated in~\cref{fig:pipeline} (top-left).

\noindent%
\textbf{Data Annotation.}
To obtain reliable human preference data, we recruited six trained annotators, each labeling about 5,664 instances.
Annotators were given basic guidelines on the task but made their rankings based on personal judgment of consistency with the reference image.
A random subset of 3,136 instances is reserved as the \textit{ConsistencyRank} benchmark for evaluating reward models

Directly training on ranking data is computationally demanding and often requires non-trivial architectural changes~\cite{wu2025editrewardhumanalignedrewardmodel}.
To retain both simplicity and generality, we convert rankings into pairwise comparisons~(see~\cref{fig:pipeline}, top-right),
preserving rich supervision while providing clear positive and negative pairs for consistency evaluation.
To further improve diversity and balance, we incorporate 5,695 manually verified consistent pairs from ShareGPT-4o-Image~\cite{chen2025sharegpt4oimagealigningmultimodalmodels}.
In total, the dataset contains 54,624 annotated image pairs (27,599 consistent and 27,025 inconsistent),
each labeled through human evaluation and augmented with CoT reasoning annotations generated by GPT-5~\cite{OpenAI2025IntroducingGPT5}.
These reasoning annotations enhance dataset interpretability and help mitigate overfitting during \textit{Paco-Reward} training.



\subsection{PaCo-Reward: Pairwise Reward Modeling}
\label{subsec:consistencyreward}
\textbf{Motivation.}
Consistency evaluation inherently requires comparing multiple images to assess their alignment on shared attributes.
Recent works~\cite{luo2025editscoreunlockingonlinerl,wu2025editrewardhumanalignedrewardmodel} demonstrate that VLM-based reward models are effective for such multi-image, text-conditioned tasks.
However, existing approaches have key limitations.
Some introduce extra regression heads to output scalar rewards, which mismatch with the next-token prediction nature of VLMs.
Others rely on long CoT reasoning to infer reward scores, causing high computational overhead during RL training.

\noindent%
\textbf{Reward Modeling.}
Inspired by prior works~\cite{jia2025settleonetexttoimagesetgeneration,xue2025dancegrpounleashinggrpovisual},
we propose \textit{PaCo-Reward}, a pairwise reward modeling framework that simplifies consistency learning.
Rather than modeling full rankings,
it learns relative preferences from pairwise comparisons to provide rich and efficient supervision.
To align with the autoregressive nature of VLMs,
PaCo-Reward further reformulates consistency evaluation as a generative task,
predicting the likelihood of ``Yes'' or ``No'' to indicate the consistency between two images.
This formulation naturally fits the autoregressive next-token prediction paradigm of VLMs
and can be further enhanced with CoT reasoning to improve interpretability and robustness, as illustrated in~\cref{fig:pipeline} (top-right).
During inference, the predicted probability of ``Yes'' serves as the consistency score between two images,
thus facilitating efficient pairwise consistency evaluation for RL training.
Moreover, human preference rankings can be inferred from the consistency scores of candidate images relative to the reference image, as shown in~\cref{fig:pipeline} (bottom-left).

\noindent%
\textbf{Training Objective.}
Training solely on binary responses can lead to overfitting and limited generalization,
while directly optimizing the full CoT sequence may weaken the main supervision signal.
To balance these factors, we design a weighted likelihood objective.
As shown in~\cref{fig:pipeline} (bottom-right),
given two images $I_A$ and $I_B$ under a shared prompt $P$ (collectively denoted as input $I=(I_A, I_B, P)$),
the model generates a binary answer (``Yes'' or ``No'') followed by a CoT-style reasoning sequence of $n-1$ token.
The objective maximizes the weighted likelihood of the ground-truth response:
\begin{equation*}\label{eq:paco_loss}
\mathcal{L}_{\text{PaCo}} = - \left[ \alpha \log p(y_0 \mid I) + \frac{(1 - \alpha)}{n-1} \sum_{i=1}^{n-1}\log p(y_i \mid I) \right],
\end{equation*}
where $y_0$ denotes the first token (``Yes'' or ``No''),
$y_i$ is the $i$-th token of the reasoning sequence,
and $\alpha \in [0,1]$ controls the balance between the decision token and reasoning supervision.
When $\alpha = \frac{1}{n}$, the formulation reduces to standard maximum likelihood estimation.
Through a hyperparameter search, we find that setting $\alpha=0.1$ achieves the best generalization performance in our setting.

\section{Paco-GRPO}
\label{sec:training_strategies}
\begin{figure*}[h]
    \centering
    \includegraphics[width=0.8\textwidth]{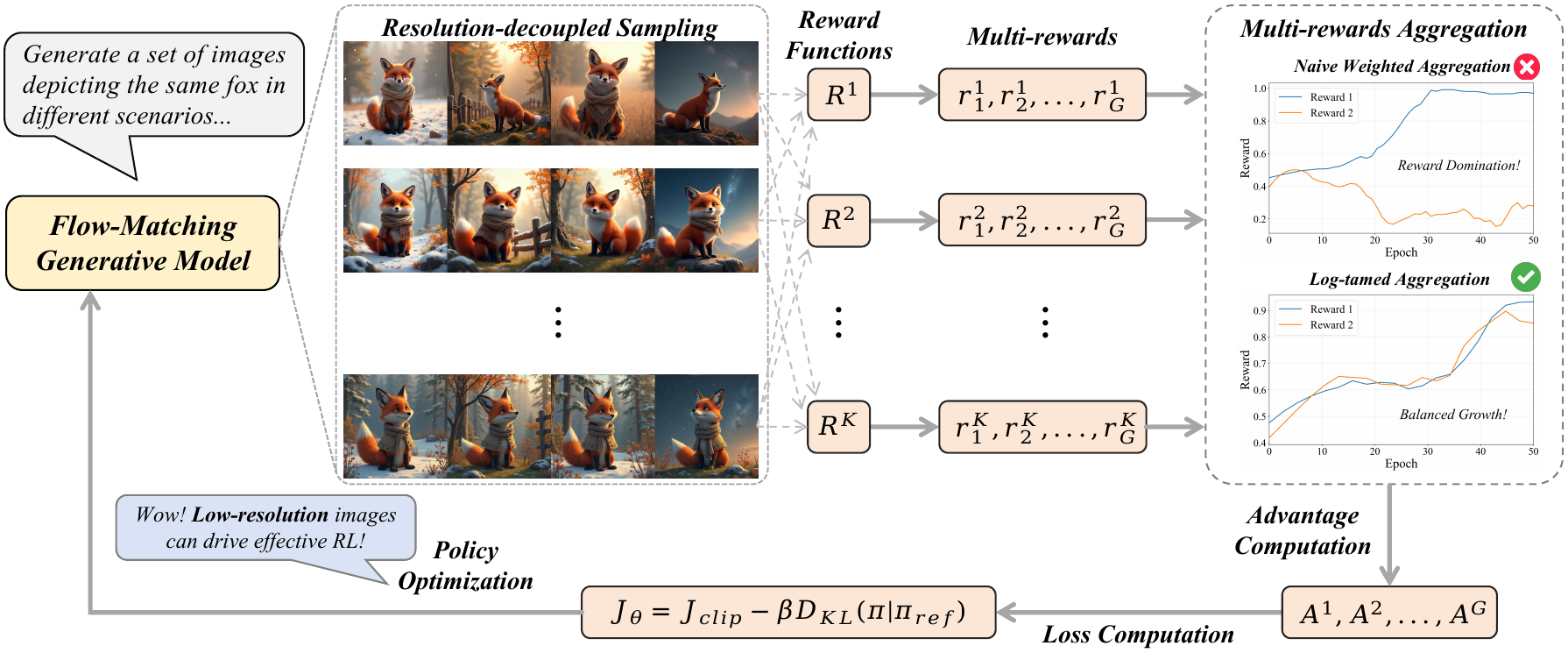}
    \vspace{-0.7em}
    \caption{Overview of our proposed \textbf{PaCo-GRPO} framework on Text-to-ImageSet generation task.
    }
    \label{fig:rl_pipeline}
    \vspace{-1.5em}
\end{figure*}

\subsection{Preliminaries}\label{subsec:preliminaries}
Recently, FlowGRPO~\cite{liu2025flowgrpotrainingflowmatching} and DanceGRPO~\cite{xue2025dancegrpounleashinggrpovisual}
have successfully adapted Group Relative Policy Optimization (GRPO)~\cite{shao2024deepseekmathpushinglimitsmathematical} to image generation models.
These methods introduce stochasticity into flow-matching models by converting deterministic ODEs into SDEs.
The corresponding SDE update rule is:
\begin{equation*}\label{eq:update_rule}
    \bm{x}_{t+\Delta t} = \bm{x}_t + \left[\bm{v}_{\theta} + \frac{\sigma_t^2}{2t}\!\left(\bm{x}_t+(1-t)\bm{v}_{\theta}\right)\right]\!\Delta t + \sigma_t\sqrt{\Delta t}\,\epsilon,
\end{equation*}
where $\bm{x}$ is the image latent, $t$ denotes the time step,
$\bm{v}_{\theta}$ is the learned velocity field for $\bm{x}_t$ at $t$,
$\epsilon\sim\mathcal{N}(0,\bm{I})$ is standard Gaussian noise,
and $\sigma_t$ is the noise scale.
This stochastic sampling promotes diversity and enhances exploration during RL training.
The GRPO objective is then maximized:
\begin{equation}\label{eq:grpo_objective}
\begin{aligned}
J_{\theta} &=  J_{\text{clip}} - \beta D_{\text{KL}}(\pi_{\theta} || \pi_{\text{ref}}),\\
J_{\text{clip}} &= \text{mean}_{i,t} \Big(
\min ( r^i_t(\theta) \hat{A}^i_t, \text{clip}\!\left( r^i_t(\theta), \varepsilon \right)\hat{A}^i_t ) \Big).
\end{aligned}
\end{equation}
where $\hat{A}^i_t$ is the estimated advantage,
$r^i_t(\theta)$ is the ratio between the current policy $\pi_{\theta}$ and the previous policy $\pi_{\theta_{\text{old}}}$,
and the KL-loss $D_{\text{KL}}$ ensures that the updated policy $\pi_{\theta}$ remains close to the reference policy $\pi_{\text{ref}}$.

Consistent image generation tasks often involve multiple images and require the simultaneous optimization of several reward signals,
such as visual consistency and prompt alignment.
This presents significant challenges in two aspects:
(i) \emph{RL efficiency}:
optimizing large or multiple images substantially increases computational cost.
Prior work~\cite{li2025mixgrpounlockingflowbasedgrpo,he2025tempflowgrpotimingmattersgrpo,liu2025flowgrpotrainingflowmatching} demonstrates that training can be reduced to 1-2 steps without compromising performance,
yet sampling efficiency, which remains the primary computational bottleneck, is largely unaddressed.
(ii) \emph{RL stability}:
multi-reward optimization is prone to reward domination~\cite{roijers2013survey,hayes2021practical},
where one signal can dominate others, resulting in suboptimal outcomes or destabilized training.

This motivates the following question:
\textbf{\textit{Can we improve both sampling and training efficiency simultaneously while ensuring stable multi-reward optimization?}}

\subsection{PaCo-GRPO Strategies}\label{subsec:paco_grpo}
To address these challenges, we propose \textit{PaCo-GRPO}, which introduces two key strategies for efficient and stable RL training:
(i) a resolution-decoupled training strategy that improves both sampling and training efficiency, and
(ii) a log-tamed multi-reward aggregation scheme that adaptively mitigates reward domination and stabilizes optimization.

\noindent%
\textbf{Resolution-decoupled Training.}\label{subsubsec:resolution_decoupled_training}
In Text-to-ImageSet generation, models often produce high-resolution outputs containing multiple sub-figures,
each matching the standard resolution of text-to-image tasks.
This requirement dramatically increases computational cost,
scaling quadratically with image resolution in Transformer-based architectures~\cite{vaswani2017attention}.
Such scaling makes RL particularly expensive.

FlowGRPO~\cite{liu2025flowgrpotrainingflowmatching} has observed that even low-quality images generated with fewer denoising steps can still provide effective reward signals for RL optimization.
Building on this insight, we introduce a \emph{resolution-decoupled training} strategy to improve efficiency.
While the model generates high-resolution ($h\times w$) images during inference and evaluation,
it produces lower-resolution images (e.g., $\frac{h}{2}\times \frac{w}{2}$) during training for reward computation and optimization.
This approach substantially reduces the overhead of both sampling and training,
thereby accelerating RL optimization (see~\cref{fig:resolution_train}).
It also integrates seamlessly with existing strategies~\cite{li2025mixgrpounlockingflowbasedgrpo,liu2025flowgrpotrainingflowmatching,he2025tempflowgrpotimingmattersgrpo} for further efficiency gains.

\noindent%
\textbf{Log-tamed Multi-reward Aggregation}\label{subsubsec:log_tamed_multi_reward_aggregation}
During RL training~\cite{schulman2017trustregionpolicyoptimization,schulman2017proximalpolicyoptimizationalgorithms,rafailov2024direct,shao2024deepseekmathpushinglimitsmathematical},
multiple reward signals are often combined to guide models toward desired behaviors.
For consistent image generation, both visual consistency and prompt alignment are essential for ensuring high-quality outputs.
Given $N$ input conditions $\{\bm{c}_i\}_{i=1}^N$,
the model generates a group of $G$ samples $\{\bm{x}_i^j\}_{j=1}^G$ for each condition $\bm{c}_i$.
Each reward model $R^k$ outputs a scalar score $R^k(\bm{x}_i^j, \bm{c}_i)$,
and the aggregated reward and advantage are computed as
\begin{equation}\label{eq:naive_multi_reward}
    \hat{r}_i^j = \sum_{k=1}^K w_k R^k(\bm{x}_i^j, \bm{c}_i),\, \hat{A}_i^j = \frac{\hat{r}_i^j - \text{mean}_j(\hat{r}_i^j)}{\text{std}_j(\hat{r}_i^j)},
\end{equation}
where $w_k$ is the weight for the $k$-th reward.
However, this naive aggregation often suffers from \textit{reward domination}~\cite{roijers2013survey,hayes2021practical},
where one reward dominates optimization, leading to suboptimal or even degenerate results.

This issue can be mitigated by manual weight tuning, but the process is labor-intensive and lacks generality.
To address this, we propose a log-tamed aggregation strategy,
which adaptively compresses rewards with large fluctuations,
thus preventing them from overwhelming the overall optimization.
Specifically, we first compute the coefficient of variation~\cite{wikipedia_coefficient_of_variation} for the $k$-th reward, denoted as $h^k$:
\begin{equation}\label{eq:reward_cv}
    h^k = \frac{\text{std}_{i,j}\!\left(R^k(\bm{x}_i^j, \bm{c}_i)\right)}{\text{mean}_{i,j}\!\left(R^k(\bm{x}_i^j, \bm{c}_i)\right)}.
\end{equation}
A high $h^k$ indicates that the $k$-th reward exhibits large fluctuations,
leading to large absolute values in the aggregated reward and potentially dominating optimization.
To temper such effects, we apply a logarithmic transformation:
\begin{equation}\label{eq:log_tamed_reward}
    \overline{R}^k(\bm{x}_i^j, \bm{c}_i) =
    \begin{cases}
    \log(1 + R^k(\bm{x}_i^j, \bm{c}_i)), & \text{if } h^k > \delta, \\
    R^k(\bm{x}_i^j, \bm{c}_i), & \text{otherwise},
    \end{cases}
\end{equation}
where $\delta$ is a threshold hyperparameter, which can be dynamically set as the mean of $\{h^k\}_{k=1}^K$ or as a fixed value (e.g., $0.2$) based on prior knowledge.
This transformation compresses large reward values while preserving the relative order of samples,
thus mitigating reward domination without distorting the underlying preferences.
\newcommand{\compositional}{}
\newcommand{\agentic}{}
\newcommand{\commercial}{}
\newcommand{\unified}{}

\section{Experiments}
\label{sec:experiments}
Our experiments are designed to address the following research questions (RQs):

\noindent%
\textbf{RQ1:} Can \textit{PaCo-Reward} better capture human preferences for visual consistency than existing methods?

\noindent%
\textbf{RQ2:} Does integrating \textit{PaCo-Reward} into RL training improve the performance of consistent image generation?

\noindent%
\textbf{RQ3:} Do the proposed \textit{PaCo-GRPO} strategies improve the efficiency and stability of RL training?

\subsection{Experimental Setup}
Our experiments consist of two stages: reward modeling evaluation and RL evaluation.

\subsubsection{Reward Modeling Evaluation Setup.}
\textbf{Models.}
We trained two variants of \textit{PaCo-Reward} based on Qwen2.5-VL-7B-Instruct~\cite{bai2025qwen25vltechnicalreport}:
(1) \textit{PaCo-Reward-7B-Fast}, trained with only binary labels for fast convergence;
(2) \textit{PaCo-Reward-7B}, trained on the full \textit{PaCo-Dataset} with reasoning-augmented labels for improved performance.

\noindent%
\textbf{Benchmarks.}
We evaluated \textit{PaCo-Reward} on two benchmarks:
(1) \textit{ConsistencyRank},
which contains $\sim$3k human-annotated instances, each containing one reference image and four candidate images ranked by visual consistency.
(2) \textit{EditReward-Bench}~\cite{wu2025editrewardhumanalignedrewardmodel}, comprising 3k pairs of original and edited images with human preference labels over Prompt Following (PF) and Consistency (C).

\noindent%
\textbf{Baselines.}
We compared \textit{PaCo-Reward} against state-of-the-art reward models, including
CLIP-I~\cite{radford2021learningtransferablevisualmodels}, DreamSim~\cite{fu2023dreamsim}, InternVL3.5-8B~\cite{wang2025internvl35advancingopensourcemultimodal}, and Qwen2.5-VL-7B~\cite{bai2025qwen25vltechnicalreport} on \textit{ConsistencyRank};
and GPT-4.1~\cite{openai2024gpt4technicalreport}, GPT-5~\cite{OpenAI2025IntroducingGPT5}, Gemini2.5-Pro~\cite{comanici2025gemini25pushingfrontier}, the Qwen2.5-VL series~\cite{bai2025qwen25vltechnicalreport}, and the EditScore series~\cite{luo2025editscoreunlockingonlinerl} on \textit{EditReward-Bench}.

\noindent%
\textbf{Metrics.}
We evaluate using Accuracy,
Kendall's rank correlation coefficient ($\tau$), Spearman's rank correlation coefficient ($\rho$), and Top-1-Bottom-1 (T1-B1) Accuracy on \textit{ConsistencyRank},
and PF, C, and Overall (geometric mean of PF and C) on \textit{EditReward-Bench}, following~\cite{luo2025editscoreunlockingonlinerl}.

\subsubsection{RL Evaluation Setup.}
\textbf{Models.}
For models, we used FLUX.1-dev~\cite{blackforestlabs_flux1dev_2024} for Text-to-ImageSet generation,
and FLUX.1-Kontext-dev~\cite{labs2025flux1kontextflowmatching} as well as Qwen-Image-Edit~\cite{wu2025qwenimagetechnicalreport} for Image Editing tasks.
PaCo-GRPO was employed to train these models using \textit{PaCo-Reward-7B} as the reward model.
The prompt templates for each task are provided in Appendix~\ref{app:prompt_templates}.

\noindent%
\textbf{Benchmarks and Metrics.}
For benchmarks, we conducted evaluations on \textit{T2IS-Bench}~\cite{jia2025settleonetexttoimagesetgeneration} for Text-to-ImageSet and \textit{GEdit-Bench}~\cite{wu2025qwenimagetechnicalreport} for Image Editing.
Evaluation metrics followed the corresponding benchmarks:
comprehensive metrics from \textit{T2IS-Bench} for quality and consistency,
and Semantic Consistency (SC), Prompt Quality (PQ), Overall scores from \textit{GEdit-Bench} for editing performance.

\begin{table}[htbp]
    \centering
    \vspace{-0.5em}
    \caption{Benchmark results on \textit{EditReward-Bench.}}
    \vspace{-1em}
    \label{tab:editreward_bench}
    \small
    \resizebox{1\columnwidth}{!}{
    \begin{tabular}{lccc}
    \toprule
    \multirow{2}{*}{Method} &\multicolumn{3}{c}{Accuracy}\\
    & Prompt Following & Consistency & Overall\\
    \midrule
    GPT-4.1 & 0.673 & 0.602 & 0.705\\
    GPT-5 & \textbf{0.777} & \underline{0.669} & \textbf{0.755}\\
    Gemini2.5-Pro & 0.703 & 0.560 & 0.722\\
    Qwen2.5-VL-7B & 0.458 & 0.325 & 0.432\\
    Qwen2.5-VL-32B & 0.498 & 0.376 & 0.563\\
    Qwen2.5-VL-72B & 0.540 & 0.435 & 0.621\\
    EditScore-7B & 0.592 & 0.591 & 0.659\\
    EditScore-32B & 0.638 & 0.556 & 0.680\\
    EditScore-72B & 0.635 & 0.586 & 0.703\\
    \rowcolor{blue!10}
    PaCo-Reward-7B-Fast & 0.748 & 0.697 & 0.728\\
    \rowcolor{blue!10}
    PaCo-Reward-7B & \textbf{0.777} & \textbf{0.709} & \underline{0.751}\\
    \bottomrule
    \end{tabular}
    }
    \vspace{-2.5em}
\end{table}

\begin{table}[htbp]
    \centering
    \caption{Benchmark results on \textit{ConsistencyRank}.}
    \vspace{-1em}
    \label{tab:consistency_rank}
    \small
    \resizebox{1\columnwidth}{!}{
    \begin{tabular}{lcccc}
    \toprule
    Method & Accuracy $\uparrow$ & $\tau$ $\uparrow$ & $\rho$ $\uparrow$ & T1-B1 $\uparrow$ \\
    \midrule
    Clip-I & 0.394 & 0.178 & 0.206 & 0.475 \\
    DreamSim & 0.403 & 0.184 & 0.214 & 0.493 \\
    InternVL3.5-8B & 0.359 & 0.149 & 0.176 & 0.420 \\
    Qwen2.5-VL-7B & 0.344 & 0.118 & 0.138 & 0.401 \\
    \rowcolor{blue!10}
    PaCo-Reward-7B-Fast & \underline{0.441} & \underline{0.240} & \underline{0.278} & \underline{0.544} \\
    \rowcolor{blue!10}
    PaCo-Reward-7B & \textbf{0.449} & \textbf{0.250} & \textbf{0.288} & \textbf{0.557} \\
    \bottomrule
    \end{tabular}
    }
    \vspace{-1em}
\end{table}

\begin{table*}[t]
    \centering
    \small
    \renewcommand\arraystretch{1.1}
    \caption{Comparisons with various Text-to-ImageSet generation methods on T2IS-Bench. %
Scores for \textbf{Visual Consistency} are evaluated by two independent evaluators, Qwen2.5-VL-7B and Gemma-3-4B (values before/after the slash), %
to ensure cross-model reliability.
}
    \vspace{-1em}
    \label{tab:t2is_benchmark_result}
    \resizebox{1\textwidth}{!}{
    \begin{tabular}{lcccccccc}
    \toprule[1.5pt]
        \multirow{2}{*}{\textbf{Method}} & \multirow{2}{*}{\textbf{Aesthetics}} & \multicolumn{3}{c}{\textbf{Prompt Alignment}} & \multicolumn{3}{c}{\textbf{Visual Consistency}} & \multirow{2}{*}{\textbf{Avg.}} \\ 
        & & Entity & Attribute & Relation & Identity & Style & Logic \\ 
        \midrule
        \multicolumn{9}{c}{\textbf{\textit{Close Source}}} \\
        \unified GPT-4o & 0.445 & 0.663 & 0.683 & 0.693 & 0.400 / 0.792 & 0.463 / 0.795 & 0.383 / 0.840 & 0.501 / 0.697 \\
        \unified Gemini 2.0 Flash & 0.430 & 0.738 & 0.747 & 0.743 & 0.428 / 0.750 & 0.383 / 0.739 & 0.392 / 0.793 & 0.509 / 0.673 \\
        \agentic AutoT2IS + GPT-4o & 0.567 & 0.754 & 0.761 & 0.763 & 0.441 / 0.806 & 0.520 / 0.825 & 0.416 / 0.860 & 0.571 / 0.754 \\
        \unified Seedream 4.0 & 0.551 & 0.813 & 0.800 & 0.805 & 0.486 / 0.809 & 0.479 / 0.763 & 0.458 / 0.862 & 0.589 / 0.758 \\
        \midrule
        \multicolumn{9}{c}{\textbf{\textit{Open Source}}} \\
        \compositional Gemini \& SD3 & 0.500 & \textbf{0.796} & \textbf{0.794} & \textbf{0.789} & 0.287 / 0.640 & 0.244 / 0.621 & 0.320 / 0.755 & 0.480 / 0.648 \\
        \compositional Gemini \& Pixart & 0.447 & 0.743 & 0.765 & 0.747 & 0.206 / 0.615 & 0.279 / 0.638 & 0.268 / 0.737 & 0.440 / 0.617\\
        \compositional Gemini \& Hunyuan & 0.410 & 0.758 & 0.774 & 0.765 & 0.197 / 0.605 & 0.276 / 0.622& 0.271 / 0.741 & 0.436 / 0.615\\
        \compositional Gemini \& FLUX.1-dev & \underline{0.533} & \underline{0.791} & \underline{0.790} & \underline{0.786} & 0.249 / 0.636 & 0.302 / 0.619 & 0.328 / 0.754 & 0.490 / 0.659 \\
        \agentic AutoT2IS & 0.520 & 0.729 & 0.756 & 0.743 & \underline{0.359} / \underline{0.723} & \underline{0.414} / \underline{0.737} & \underline{0.356} / \underline{0.794} & \underline{0.515} / \underline{0.686}\\
        \rowcolor{blue!10}
        \unified FLUX.1-dev + PaCo-Reward-7B & \textbf{0.555} & 0.721 & 0.732 & 0.731 & \textbf{0.508} / \textbf{0.837} &  \textbf{0.549} / \textbf{0.867} & \textbf{0.422} / \textbf{0.859} & \textbf{0.576} / \textbf{0.757} \\
        \bottomrule[1.5pt]
    \end{tabular}
    }
    \vspace{-1em}
\end{table*}

\begin{table*}[htbp]
    \centering
    \caption{Benchmark results on GEdit-Bench. EN-I and EN denote English instructions, while CN-I and CN denote Chinese instructions.
    }
    \vspace{-1em}
    \label{tab:gedit_bench_result}
    \resizebox{1.0\textwidth}{!}{
    \begin{tabular}{lcccccccccccc}
    \toprule
    \multirow{2}{*}{Method} &\multicolumn{3}{c}{EN-I}&\multicolumn{3}{c}{EN} &\multicolumn{3}{c}{CN-I} &\multicolumn{3}{c}{CN}\\
    & SC & PQ & Overall & SC & PQ & Overall & SC & PQ & Overall & SC & PQ & Overall\\
    \midrule
    \multicolumn{12}{c}{\textbf{\textit{Close Source}}} \\
    SeedEdit3.0~\cite{wang2025seededit30fasthighquality} & 7.396 & 7.899 & 7.137 & 7.222 & 7.885 & 6.983 & 7.370 & 7.870 & 7.105 & 7.195 & 7.851 & 6.942\\
    GPT-Image-1~\cite{openai_gptimage1} & 7.867 & 8.097 & 7.590 & 7.743 & 8.133 & 7.494 & 7.840 & 8.075 & 7.560 & 7.708 & 8.095 & 7.451\\
    \hline
    \multicolumn{12}{c}{\textbf{\textit{Open Source}}} \\
    OmniGen~\cite{xiao2024omnigenunifiedimagegeneration} & 6.037 & 5.856 & 5.154 & 5.879 & 5.871 & 5.005 & 6.015 & 5.830 & 5.122 & 5.850 & 5.845 & 4.976\\
    OmniGen2~\cite{wu2025omnigen2explorationadvancedmultimodal} & - & - & - & 6.72 & 7.20 & 6.28 & - & - & - & - & - & -\\
    OmniGen2 + EditScore~\cite{luo2025editscoreunlockingonlinerl} & - & - & - & 7.20 & 7.46 & 6.68 & - & - & - & - & - & -\\
    Step1X-Edit~\cite{wu2025editrewardhumanalignedrewardmodel} & 7.289 & 6.962 & 6.618 & 7.131 & 6.998 & 6.444 & 7.464 & 7.076 & 6.779 & 7.647 & 7.398 & 6.983\\
    Step1X-Edit + EditReward~\cite{wu2025editrewardhumanalignedrewardmodel} & \textbf{7.895} & 6.946 & 7.131 & \textbf{7.854} & 6.931 & 7.086 & 7.757 & 7.024 & 7.074 & 7.658 & 6.995 & 7.001\\
    FLUX.1-Kontext~\cite{labs2025flux1kontextflowmatching} & 6.842 & 6.888 & 6.143 & 6.599 & 6.873 & 5.956 & - & - & - & - & - & -\\
    \rowcolor{blue!10}
    FLUX.1-Kontext + PaCo-Reward-7B & 7.279 & 7.036 & 6.636 & 7.033 & 7.088 & 6.469 & - & - & - & - & - & -\\
    Qwen-Image-Edit~\cite{wu2025qwenimagetechnicalreport} & 7.746 & 7.805 & 7.307 & 7.642 & 7.807 & 7.223 & 7.727 & 7.977 & 7.344 & 7.633 & 7.938 & 7.264\\
    \rowcolor{blue!10}
    Qwen-Image-Edit + PaCo-Reward-7B & 7.799 & \textbf{8.053} & \textbf{7.451} & 7.693 & \textbf{7.987} & \textbf{7.325} & \textbf{7.866} & \textbf{8.125} & \textbf{7.520} & \textbf{7.678} & \textbf{8.093} & \textbf{7.366}\\
    \bottomrule
    \end{tabular}}
    \vspace{-1.0em}
\end{table*}

\begin{figure*}[!h]
    \centering
    \begin{minipage}{0.3\linewidth}
        \includegraphics[width=1\linewidth]{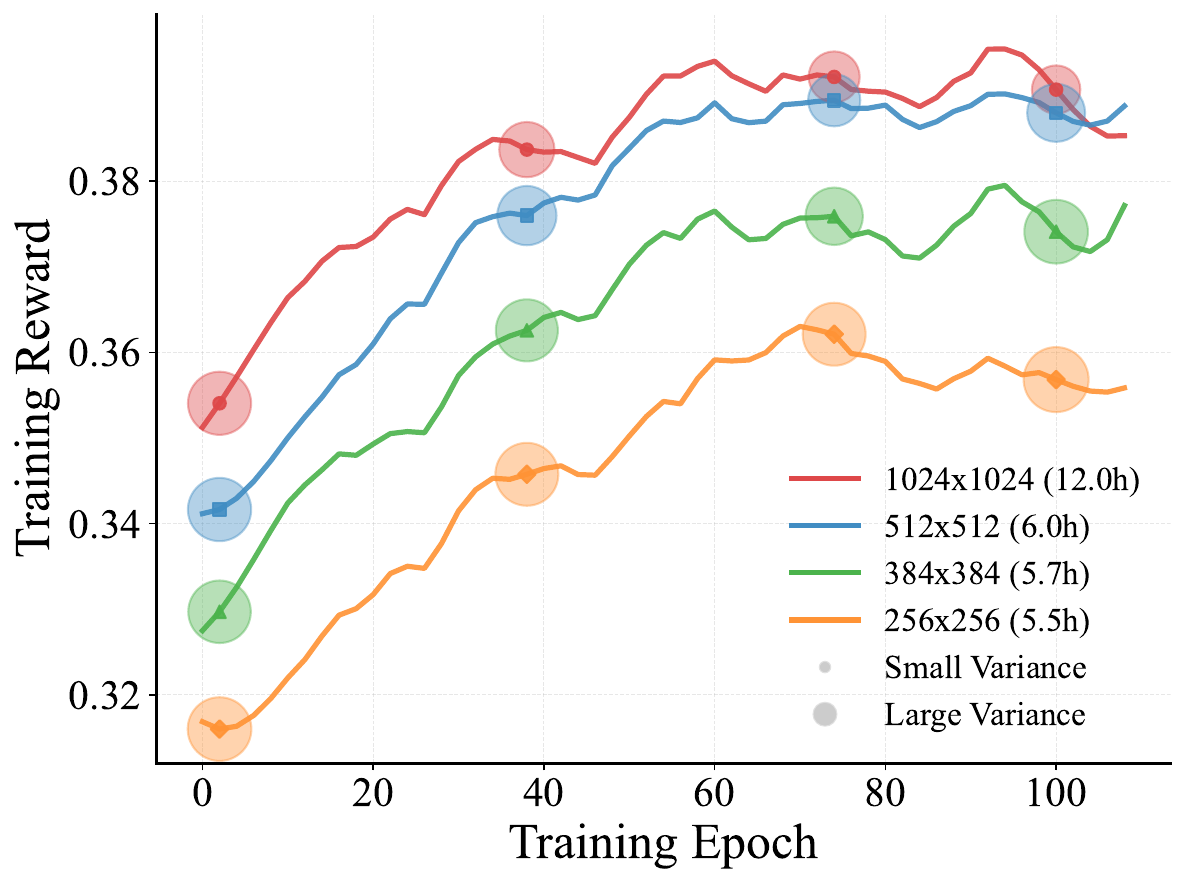}
        \vspace{-2em}
        \caption{Training processes under different image resolutions.}
        \label{fig:resolution_train}
    \end{minipage}
    \hspace{0.5em}
    \begin{minipage}{0.3\linewidth}
        \includegraphics[width=1\linewidth]{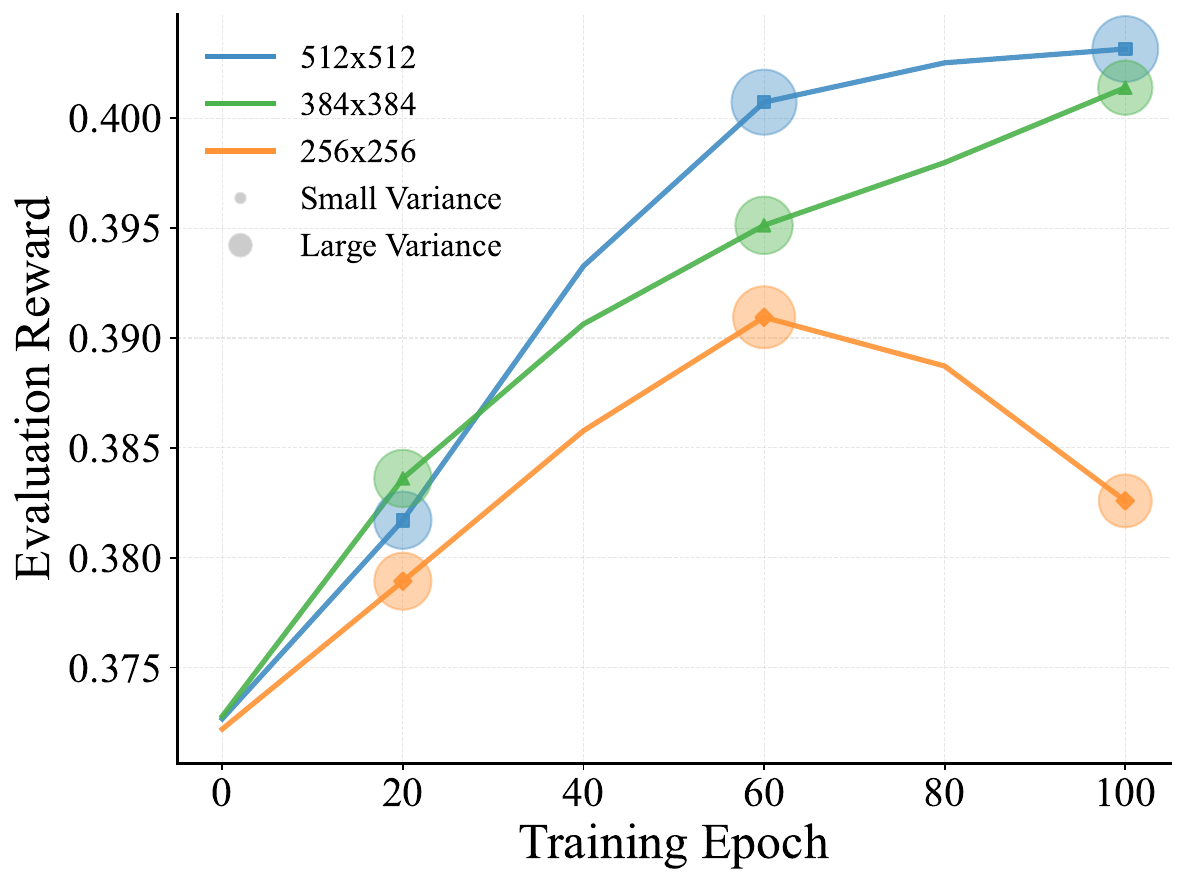}
        \vspace{-2em}
        \caption{Evaluation processes under different training image resolutions.}
        \label{fig:resolution_eval}
    \end{minipage}
    \hspace{0.5em}
    \begin{minipage}{0.3\linewidth}
        \includegraphics[width=1\linewidth]{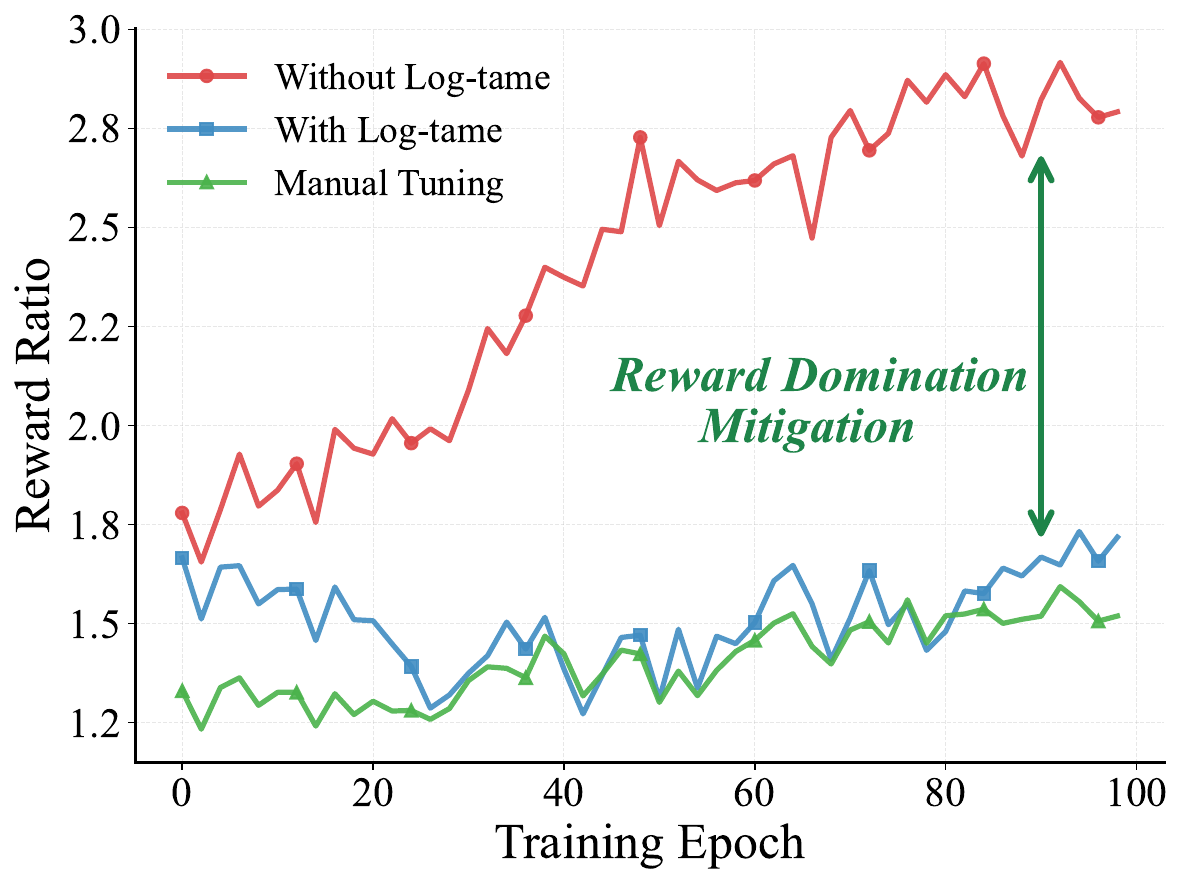}
        \vspace{-2em}
        \caption{Ablation of log-tamed aggregation on the reward ratio.}
        \label{fig:log_tame_ablation}
    \end{minipage}
    \vspace{-1.5em}
\end{figure*}

\begin{figure*}[htbp]
    \begin{minipage}{0.295\textwidth}
    \centering
    \begin{tcolorbox}
    [colback=blue!5!white,
    colframe=blue!60,
    fontupper=\fontsize{6}{8}\selectfont,
    coltitle=white,
    halign title=center,
    toptitle=1pt,
    bottomtitle=1pt,
    left=0pt,
    right=0pt,
    top=0pt,
    bottom=0pt,
    width=0.85\linewidth,boxrule=0.5pt
    ]
    \textit{
    Generate four images depicting the same dentist in scrubs in various medical scenarios.
    }
    \end{tcolorbox}
    \end{minipage}
    \begin{minipage}{0.40\textwidth}
    \centering
    \begin{tcolorbox}
    [colback=blue!5!white,
    colframe=blue!60,
    fontupper=\fontsize{6}{8}\selectfont,
    coltitle=white,
    halign title=center,
    toptitle=1pt,
    bottomtitle=1pt,
    left=0pt,
    right=0pt,
    top=0pt,
    bottom=0pt,
    width=1\linewidth,boxrule=0.5pt
    ]
    \textit{
    Create four café menu displays with chalkboard font, featuring the following headings:
    ``Fresh Brew'',``Daily Specials'', ``Homemade'' and ``Sweet Treats''.
    }
    \end{tcolorbox}
    \end{minipage}
    \begin{minipage}{0.295\textwidth}
    \centering
    \begin{tcolorbox}
    [colback=blue!5!white,
    colframe=blue!60,
    fontupper=\fontsize{6}{8}\selectfont,
    coltitle=white,
    halign title=center,
    toptitle=1pt,
    bottomtitle=1pt,
    left=0pt,
    right=0pt,
    top=0pt,
    bottom=0pt,
    width=0.9\linewidth,boxrule=0.5pt
    ]
    \textit{
    Generate four images depicting a progressive pencil drawing sequence of a young woman's portrait.
    }
    \end{tcolorbox}
    \end{minipage}
    \\
    \centering
    \includegraphics[width=\linewidth]{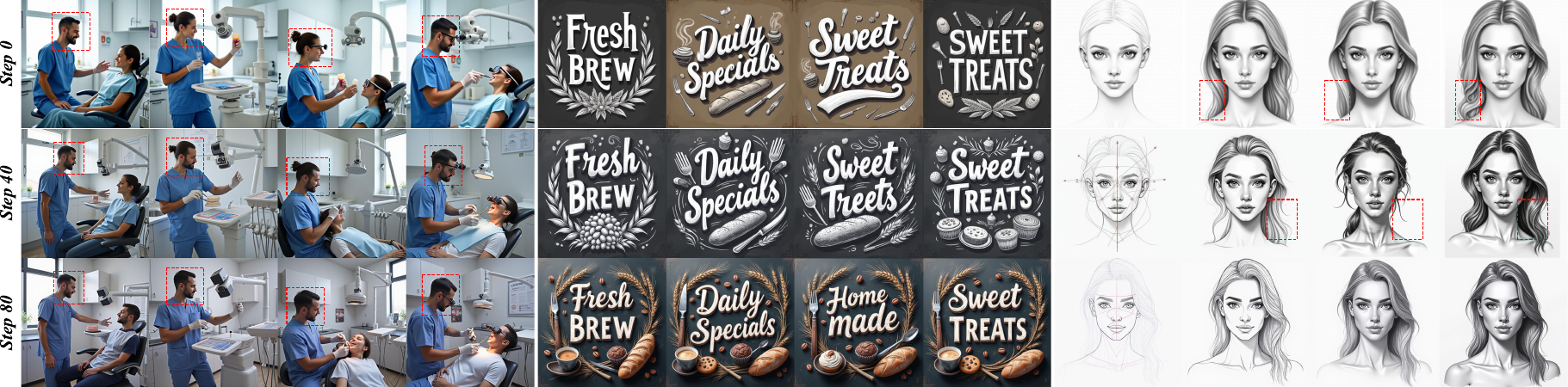}
    \vspace{-2em}
    \caption{Progressive improvement of images generated with a fixed seed during PaCo-RL training for Text-to-ImageSet generation.}
    \label{fig:training_process_grid}
    \vspace{-1.5em}
\end{figure*}



\subsection{Experimental Results}
\textbf{RA1: As shown in~\cref{tab:editreward_bench} and~\cref{tab:consistency_rank}, PaCo-Reward effectively captures human preferences for visual consistency.}
On ConsistencyRank, advanced MLLMs such as InternVL3.5-8B and Qwen2.5-VL-7B perform worse than traditional similarity-based methods like CLIP-I and DreamSim, suggesting that current MLLMs remain misaligned with human perception of visual consistency and underscoring the need for specialized reward modeling.
After fine-tuning on our PaCo-Dataset, both variants of PaCo-Reward consistently surpass all baselines.
Notably, PaCo-Reward-7B achieves a 10.5\% gain in Accuracy and a 0.150 increase in Spearman's $\rho$ over the original Qwen2.5-VL-7B, highlighting the effectiveness of our pairwise reward modeling paradigm in aligning with human preferences.
Benefiting from the diverse and high-quality supervision provided by PaCo-Dataset, PaCo-Reward-7B attains highly competitive results on EditReward-Bench, outperforming all open-source baselines and approaching the performance of proprietary models such as GPT-5.
These results demonstrate that PaCo-Reward generalizes effectively in modeling human preferences for diverse visual consistency.

\noindent%
\textbf{RA2: PaCo-Reward improves consistency in RL-based image generation.}
We integrate PaCo-Reward-7B into PaCo-GRPO for both Text-to-ImageSet generation and Image Editing tasks, as shown in~\cref{tab:t2is_benchmark_result,tab:gedit_bench_result}.
To ensure cross-model evaluation of visual consistency on T2IS-Bench,
we additionally employ Gemma-3-4B-IT~\cite{gemmateam2025gemma3technicalreport},
since PaCo-Reward-7B is fine-tuned from Qwen2.5-VL-7B.

On T2IS-Bench, FLUX.1-dev enhanced with PaCo-Reward-7B substantially outperforms the strongest open-source baseline~\cite{jia2025settleonetexttoimagesetgeneration} across all visual consistency metrics,
achieving average absolute gains of 0.117 under the Qwen2.5-VL-7B evaluator and 0.103 under the Gemma-3-4B-IT evaluator.
Moreover, it reaches performance comparable to closed-source models such as GPT-4, narrowing the gap between open-source and closed-source solutions.

On GEdit-Bench, PaCo-Reward-7B consistently improves SC and PQ across all base models.
Unlike EditReward~\cite{wu2025editrewardhumanalignedrewardmodel}, which degrades perceptual quality for Step1X-Edit,
our method achieves a balanced improvement, enhancing consistency while preserving image quality.
Even for strong baselines such as Qwen-Image-Edit, PaCo-Reward-7B further boosts both consistency and perceptual quality.
These gains stem from the strong reward signal provided by PaCo-Reward-7B,
which guides RL optimization toward consistent yet visually coherent outputs.

\noindent%
\textbf{RA3: PaCo-GRPO improves RL training efficiency and stability.}
To evaluate PaCo-GRPO's effectiveness,
we conduct ablation studies on its two key components using the challenging Text-to-ImageSet task.

\emph{Resolution-decoupled training}:
\cref{fig:resolution_train} illustrates training under different image resolutions.
Each training image uses a $2\times2$ grid layout and is divided into four sub-images for reward computation,
effectively halving the resolution in each dimension.
As shown in~\cref{fig:resolution_train},
low-resolution training ($512\times512$) starts with lower rewards but reaches the performance of $1024\times1024$ training after about 50 epochs,
demonstrating that lower-quality images still provide reliable reward feedback~\cite{liu2025flowgrpotrainingflowmatching}.
Low-resolution training also shows higher reward variance (indicated by circles in~\cref{fig:resolution_train,fig:resolution_eval}),
which promotes exploration and produces more diverse samples that enhance RL optimization.
However, as shown in~\cref{fig:resolution_eval}, $256\times256$ training fails due to insufficient visual detail for reliable reward evaluation.
These results indicate that moderate resolution reduction can greatly accelerate training without degrading final performance.

\emph{Log-tamed multi-reward aggregation}:
To assess the effect of log-tamed aggregation in mitigating reward domination,
we compare it with standard weighted-sum aggregation (see~\cref{fig:log_tame_ablation}).
Two reward components are used: \textit{Consistency} (PaCo-Reward-7B) and \textit{Prompt Alignment} (CLIP-T~\cite{radford2021learningtransferablevisualmodels}).
The plot shows the ratio of \textit{Consistency} to \textit{Prompt Alignment} rewards during training.
Under standard aggregation without careful weight tuning, this ratio quickly exceeds 2.5 after 50 epochs,
indicating that the \textit{Consistency} reward dominates optimization.
In contrast, log-tamed aggregation keeps the ratio below 1.8 throughout training,
preventing any single reward from dominating and producing balanced improvements across all objectives.

\subsection{Case Studies}
\cref{fig:training_process_grid} visualizes progressive improvements achieved during PaCo-RL training with a fixed random seed.
In the \textbf{identity} case, the dentist's facial and hairstyle attributes gradually converge toward a consistent appearance across different medical scenarios.
In the \textbf{style} case, the chalkboard font across café menus becomes increasingly uniform, with higher prompt fidelity and improved overall aesthetics.
In the \textbf{logic} case, the model learns a correct sketch-to-drawing progression, where each subsequent panel extends and refines the previous one rather than altering it.
Together, these cases demonstrate PaCo-RL's capability to jointly enhance diverse dimensions of visual consistency, validating effectiveness of both reward model and RL framework.
\section{Conclusion}
\label{sec:conclusion}
Our work introduces PaCo-RL, a novel reinforcement learning framework designed for consistent image generation.
It consists of PaCo-Reward and PaCo-GRPO, which together enable more human-aligned and computationally efficient training for consistent image generation tasks.
\section*{Acknowledgment}
\label{sec:ack}
This work is supported by the Fundamental and Interdisciplinary Disciplines Breakthrough Plan of the Ministry of Education of China (No. JYB2025XDXM101), the National Natural Science Foundation of China (No. 62272374, No. 62192781), the Natural Science Foundation of Shaanxi Province (No.2024JC-JCQN-62), the State Key Laboratory of Communication Content Cognition under Grant No. A202502, the Key Research and Development Project in Shaanxi Province (No. 2023GXLH-024)
\textbf{and}
by A*STAR Career Development Fund <Project No. C243512010>.
{
    \small
    \bibliographystyle{ieeenat_shortname}
    \bibliography{main}
}

\begin{figure*}[!hb]
\centering
\begin{minipage}[b]{0.162\textwidth}
    \centering
    \large\textbf{Ref. Image}
\end{minipage}%
\hspace{0.02\textwidth}%
\begin{minipage}[b]{0.81\textwidth}
    \centering
    \begin{tikzpicture}
        \draw[-{Stealth[length=3mm, width=3mm]}, line width=2pt] 
            (0,0) -- (0.9\textwidth,0) 
            node[midway, above, font=\bfseries\large] 
            {\textit{``Make him look stronger''}};
    \end{tikzpicture}
\end{minipage}

\begin{minipage}[t]{0.162\textwidth}
    \centering
    \includegraphics[width=\textwidth]{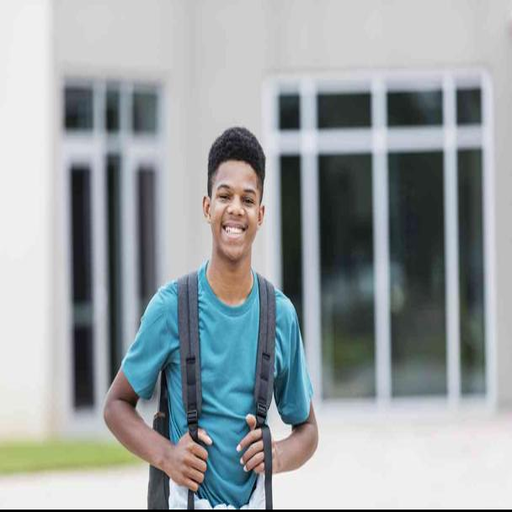}
\end{minipage}%
\hspace{0.02\textwidth}%
\begin{minipage}[t]{0.81\textwidth}
    \centering
    \includegraphics[width=\textwidth]{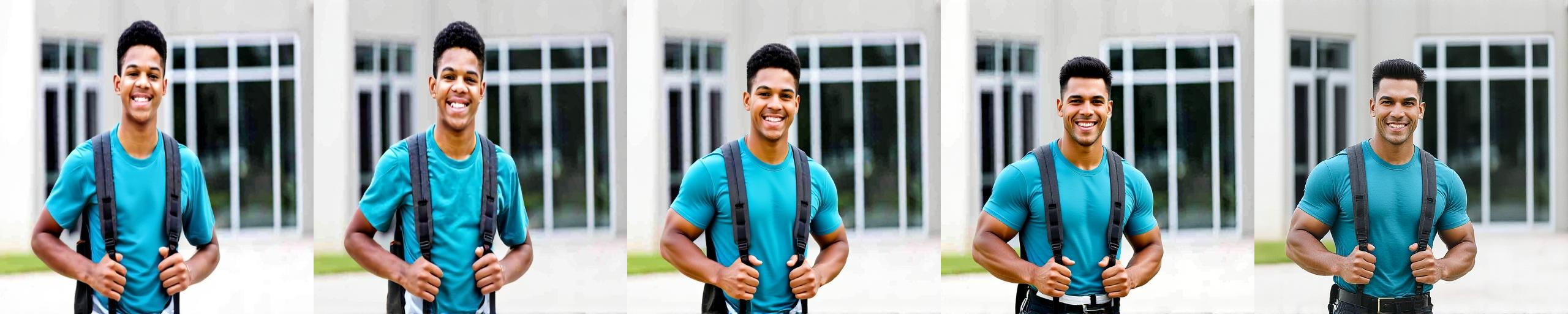}
\end{minipage}
\begin{minipage}[b]{0.162\textwidth}
    \centering
    \large\textbf{Ref. Image}
\end{minipage}%
\hspace{0.02\textwidth}%
\begin{minipage}[b]{0.81\textwidth}
    \centering
    \begin{tikzpicture}
        \draw[-{Stealth[length=3mm, width=3mm]}, line width=2pt] 
            (0,0) -- (0.9\textwidth,0) 
            node[midway, above, font=\bfseries\large] 
            {\textit{``Make the action of the woman to laughing''}};
    \end{tikzpicture}
\end{minipage}

\begin{minipage}[t]{0.162\textwidth}
    \centering
    \includegraphics[width=\textwidth]{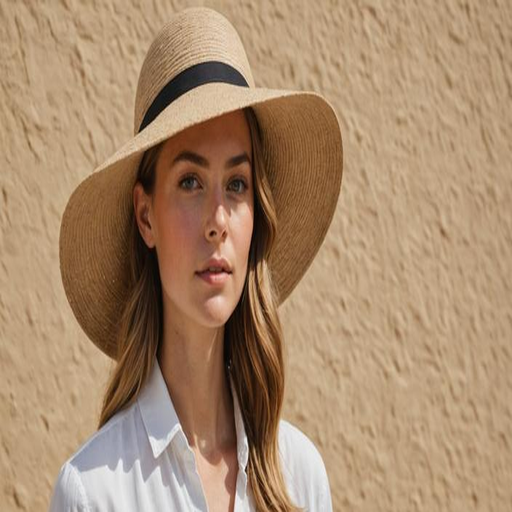}
\end{minipage}%
\hspace{0.02\textwidth}%
\begin{minipage}[t]{0.81\textwidth}
    \centering
    \includegraphics[width=\textwidth]{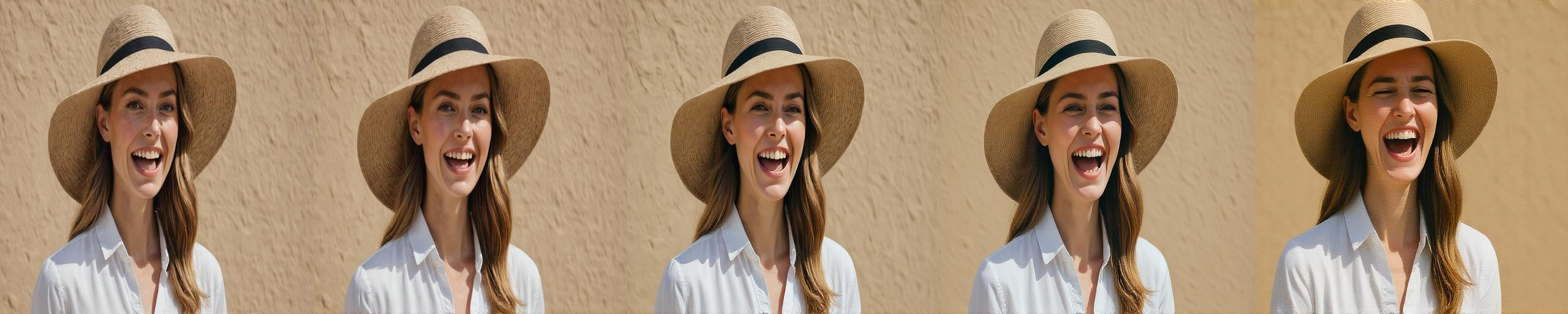}
\end{minipage}

\begin{minipage}[b]{0.162\textwidth}
    \centering
    \large\textbf{Ref. Image}
\end{minipage}%
\hspace{0.02\textwidth}%
\begin{minipage}[b]{0.81\textwidth}
    \centering
    \begin{tikzpicture}
        \draw[-{Stealth[length=3mm, width=3mm]}, line width=2pt] 
            (0,0) -- (0.9\textwidth,0) 
            node[midway, above, font=\bfseries\large] 
            {\textit{``Make him gain 20 pounds''}};
    \end{tikzpicture}
\end{minipage}

\begin{minipage}[t]{0.162\textwidth}
    \centering
    \includegraphics[width=\textwidth]{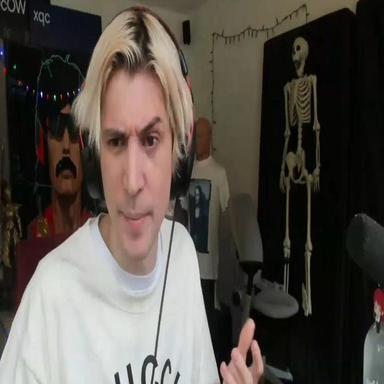}
\end{minipage}%
\hspace{0.02\textwidth}%
\begin{minipage}[t]{0.81\textwidth}
    \centering
    \includegraphics[width=\textwidth]{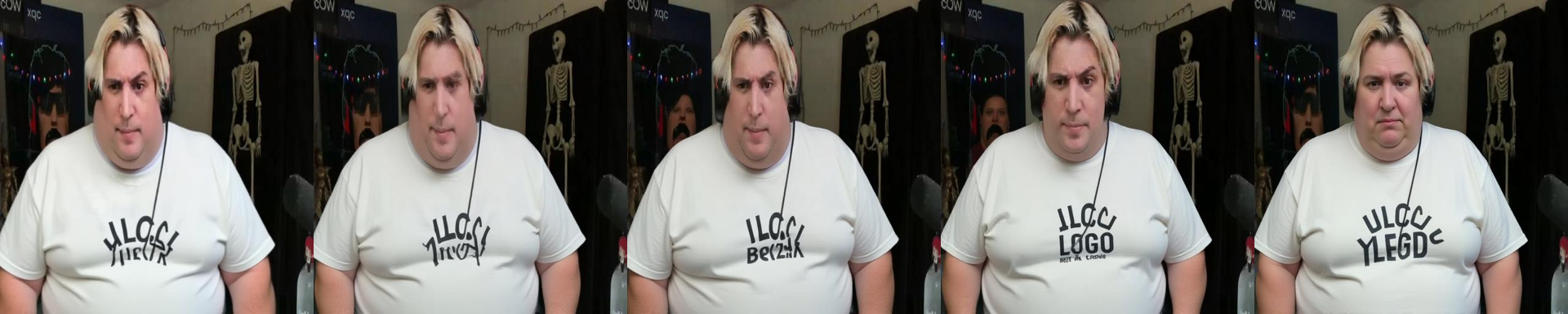}
\end{minipage}
\begin{minipage}[b]{0.162\textwidth}
    \centering
    \large\textbf{Ref. Image}
\end{minipage}%
\hspace{0.02\textwidth}%
\begin{minipage}[b]{0.81\textwidth}
    \centering
    \begin{tikzpicture}
        \draw[-{Stealth[length=3mm, width=3mm]}, line width=2pt] 
            (0,0) -- (0.9\textwidth,0) 
            node[midway, above, font=\bfseries\large] 
            {\textit{``Add abs to the original photo''}};
    \end{tikzpicture}
\end{minipage}

\begin{minipage}[t]{0.162\textwidth}
    \centering
    \includegraphics[width=\textwidth]{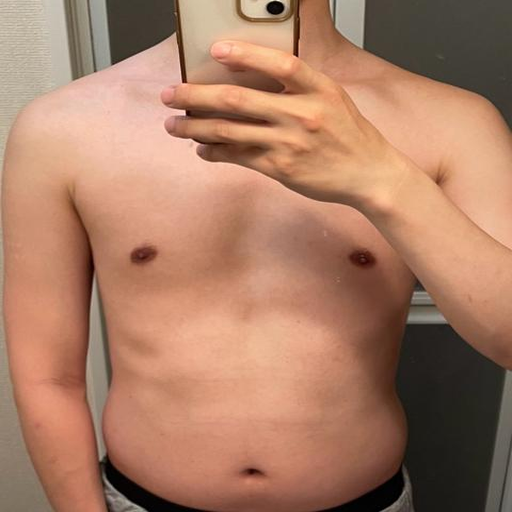}
\end{minipage}%
\hspace{0.02\textwidth}%
\begin{minipage}[t]{0.81\textwidth}
    \centering
    \includegraphics[width=\textwidth]{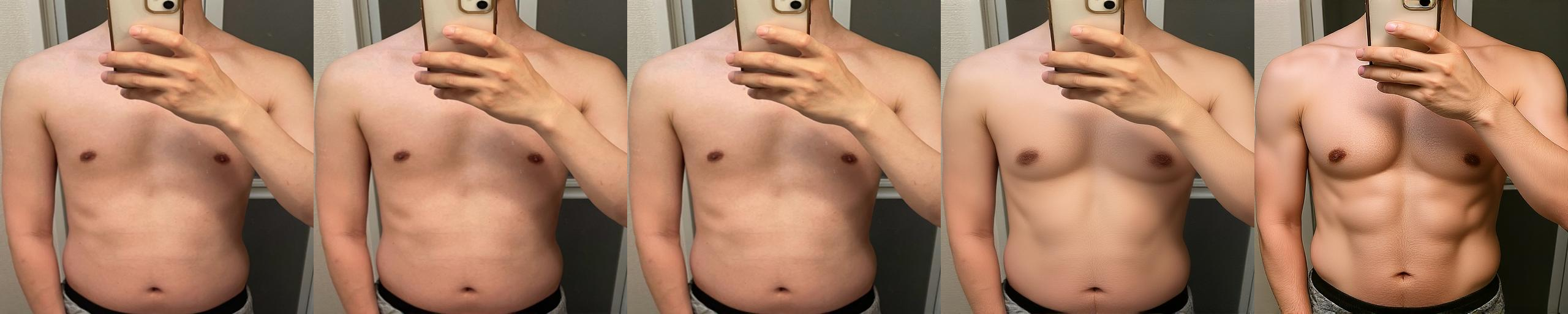}
\end{minipage}

\caption{Training progression visualization for \textit{Image Editing} with different prompts.}
\label{fig:editing_training_process}
\end{figure*}

\begin{figure*}[!t]
\centering
\begin{minipage}[c]{0.06\textwidth}
    \centering
    \begin{tikzpicture}
        \draw[-{Stealth[length=3mm, width=3mm]}, line width=2pt] (0,0.6\textheight)--(0,0)
        node[midway, right, font=\bfseries\large, rotate=90, anchor=south] 
        {Training progression};
    \end{tikzpicture}
\end{minipage}%
\hspace{0.01\textwidth}%
\begin{minipage}[c]{0.9\textwidth}
    \centering
    \begin{tcolorbox}[colback=gray!10, colframe=black, boxrule=1pt, arc=2mm, 
                      left=5pt, right=5pt, top=3pt, bottom=3pt, width=\textwidth]
        \centering
        \textbf{Prompt:} \small\textit{``Depict the chronological decomposition of a single leaf on a forest floor. All images maintain a realistic style with consistent lighting and environmental elements, focusing on the gradual transformation of the leaf while adhering to natural decay processes. The forest floor setting includes subtle elements like soil texture, scattered debris, and occasional fungi or insects.''}
    \end{tcolorbox}        
    \includegraphics[width=\textwidth]{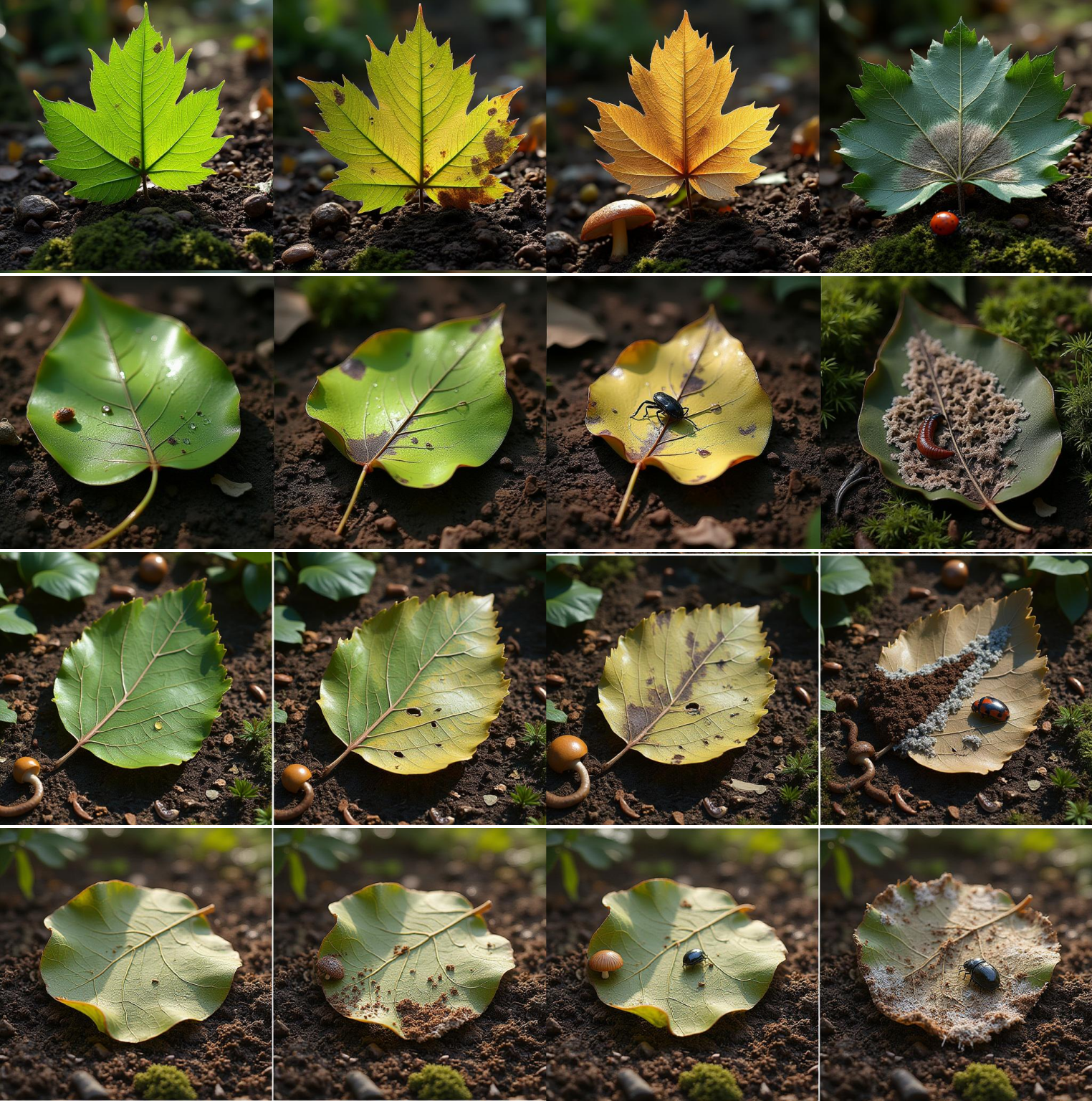}
\end{minipage}
\caption{Training progression visualization for \textit{Text-to-ImageSet} generation.}
\label{fig:t2is_training_process_1}
\end{figure*}

\begin{figure*}[!t]
\centering
\begin{minipage}[c]{0.06\textwidth}
    \centering
    \begin{tikzpicture}
        \draw[-{Stealth[length=3mm, width=3mm]}, line width=2pt] (0,0.6\textheight)--(0,0)
        node[midway, right, font=\bfseries\large, rotate=90, anchor=south] 
        {Training progression};
    \end{tikzpicture}
\end{minipage}%
\hspace{0.01\textwidth}%
\begin{minipage}[c]{0.9\textwidth}
    \centering
    \begin{tcolorbox}[colback=gray!10, colframe=black, boxrule=1pt, arc=2mm, 
                      left=5pt, right=5pt, top=3pt, bottom=3pt, width=\textwidth]
        \centering
        \textbf{Prompt:} \small\textit{``Health beverage labels featuring honey drip font with viscous liquid texture and hexagonal comb patterns. All labels utilize the honey drip font style, integrating hexagonal comb motifs and natural/organic themes. Consistency in color palette (golden, amber, earthy tones) and texture emphasis ensures visual harmony across the set.''}
    \end{tcolorbox}        
    \includegraphics[width=\textwidth]{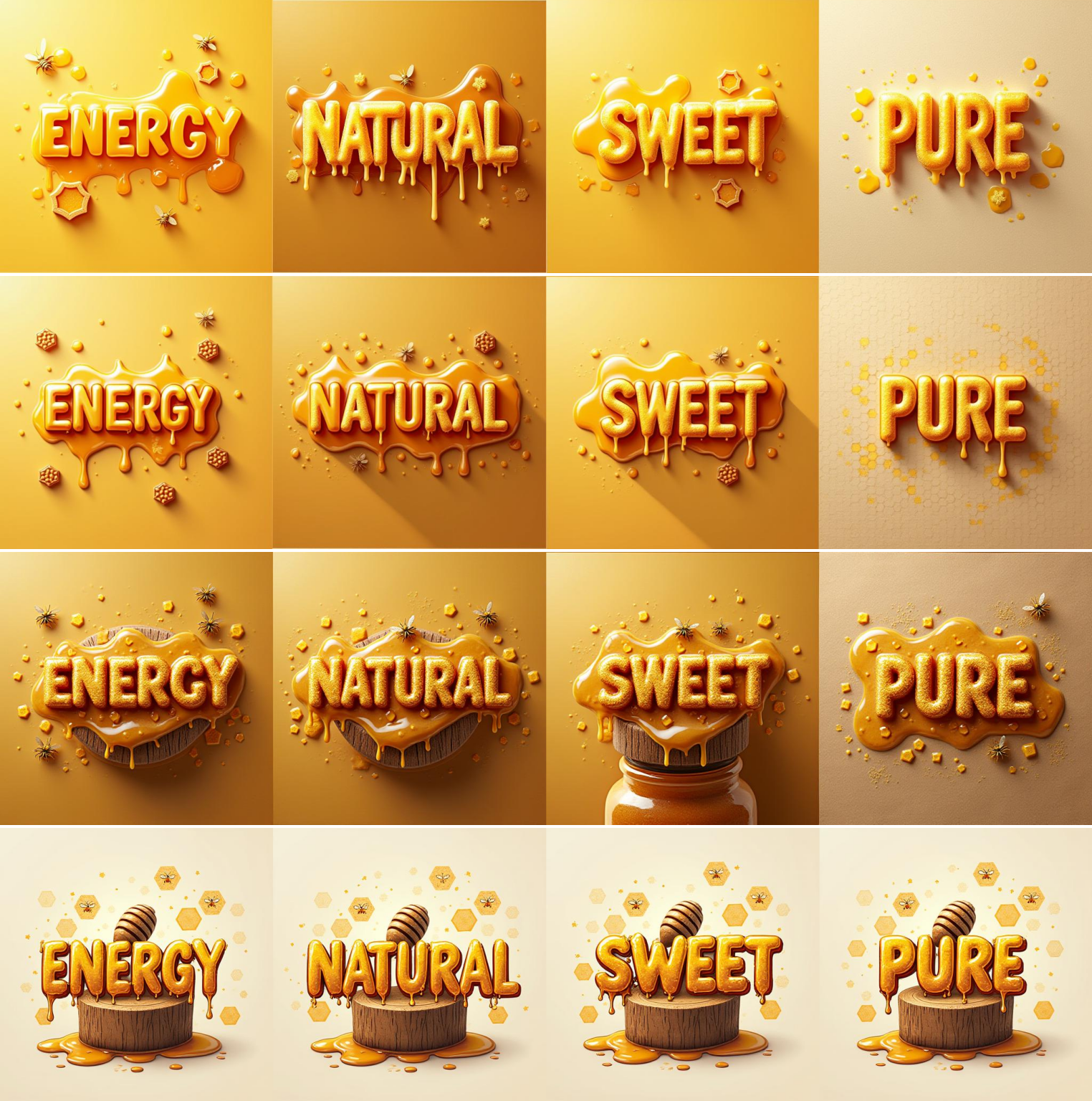}
\end{minipage}
\caption{Training progression visualization for \textit{Text-to-ImageSet} generation.}
\label{fig:t2is_training_process_2}
\end{figure*}

\begin{figure*}[!t]
\centering
\begin{minipage}[c]{0.06\textwidth}
    \centering
    \begin{tikzpicture}
        \draw[-{Stealth[length=3mm, width=3mm]}, line width=2pt] (0,0.6\textheight)--(0,0)
        node[midway, right, font=\bfseries\large, rotate=90, anchor=south] 
        {Training progression};
    \end{tikzpicture}
\end{minipage}%
\hspace{0.01\textwidth}%
\begin{minipage}[c]{0.9\textwidth}
    \centering
    \begin{tcolorbox}[colback=gray!10, colframe=black, boxrule=1pt, arc=2mm, 
                      left=5pt, right=5pt, top=3pt, bottom=3pt, width=\textwidth]
        \centering
        \textbf{Prompt:} \small\textit{``Step-by-step progression of creating a cheerful chef emoji. All images use a minimalist, cartoonish style with a clean white background. Bright and cohesive color schemes unify the stages, maintaining continuity in character proportions and playful energy.''}
    \end{tcolorbox}        
    \includegraphics[width=\textwidth]{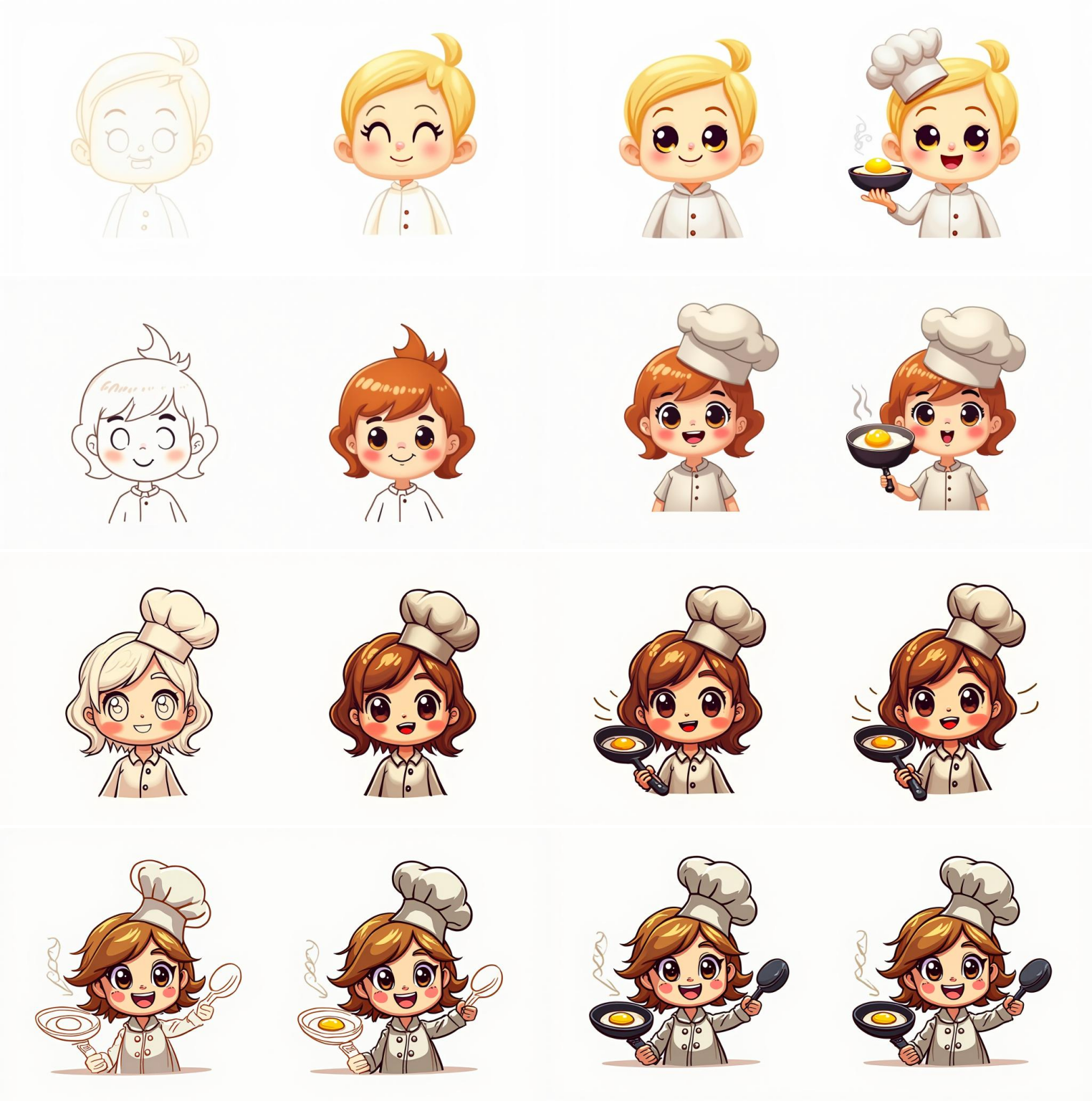}
\end{minipage}
\caption{Training progression visualization for \textit{Text-to-ImageSet} generation.}
\label{fig:t2is_training_process_3}
\end{figure*}

\begin{figure*}[!hb]
\centering

\begin{tcolorbox}[colback=gray!10, colframe=black, boxrule=1pt, arc=2mm, 
                  left=5pt, right=5pt, top=3pt, bottom=3pt, width=0.95\textwidth]
    \centering
    \textbf{Prompt:} \small\textit{``Please generate four different perspective images of a 3D animated parrot with a vibrant and colorful plumage.  The parrot exhibits a stunning array of colors, including shades of red, green, blue, and yellow, with detailed feather textures that reflect light and give a sense of depth...''}
\end{tcolorbox}

\begin{minipage}[c]{0.08\textwidth}
    \centering
    \rotatebox{90}{\large\textbf{Ours}}
\end{minipage}%
\hspace{0.01\textwidth}%
\begin{minipage}[c]{0.89\textwidth}
    \centering
    \includegraphics[width=\textwidth]{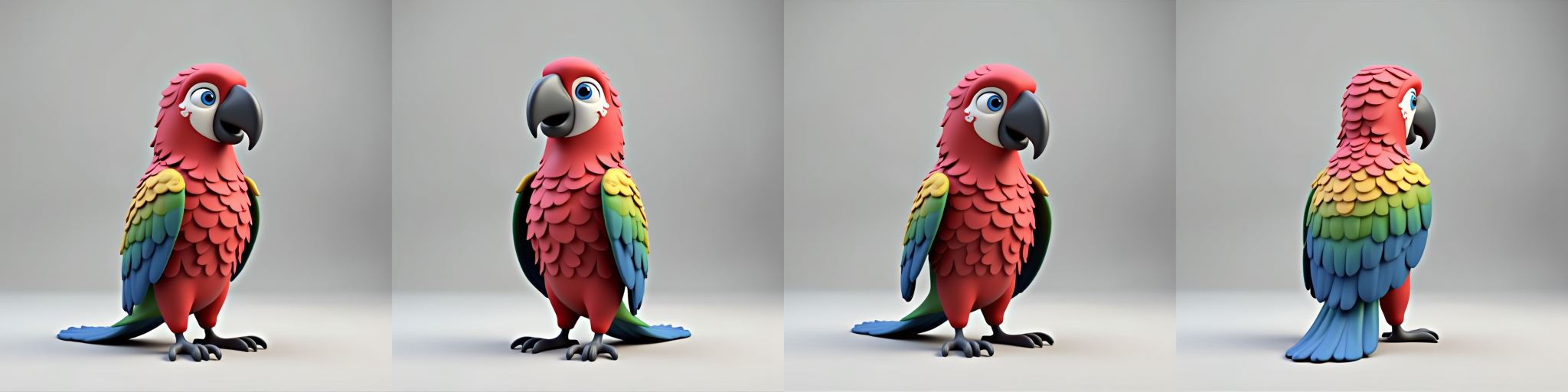}
\end{minipage}

\begin{minipage}[c]{0.08\textwidth}
    \centering
    \rotatebox{90}{\large\textbf{AutoT2IS}}
\end{minipage}%
\hspace{0.01\textwidth}%
\begin{minipage}[c]{0.89\textwidth}
    \centering
    \includegraphics[width=\textwidth]{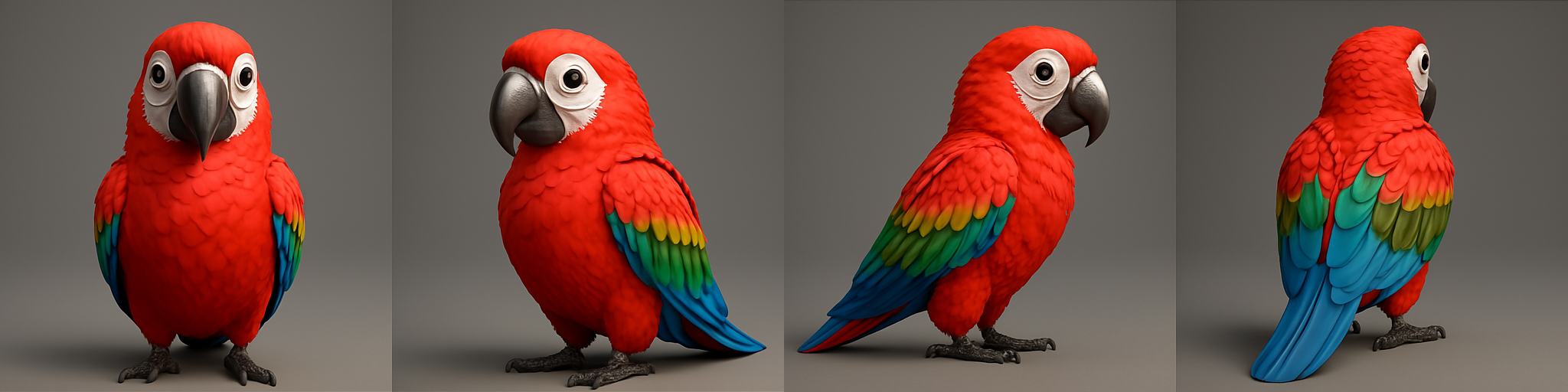}
\end{minipage}

\begin{minipage}[c]{0.08\textwidth}
    \centering
    \rotatebox{90}{\large\textbf{Seedream}}
\end{minipage}%
\hspace{0.01\textwidth}%
\begin{minipage}[c]{0.89\textwidth}
    \centering
    \includegraphics[width=\textwidth]{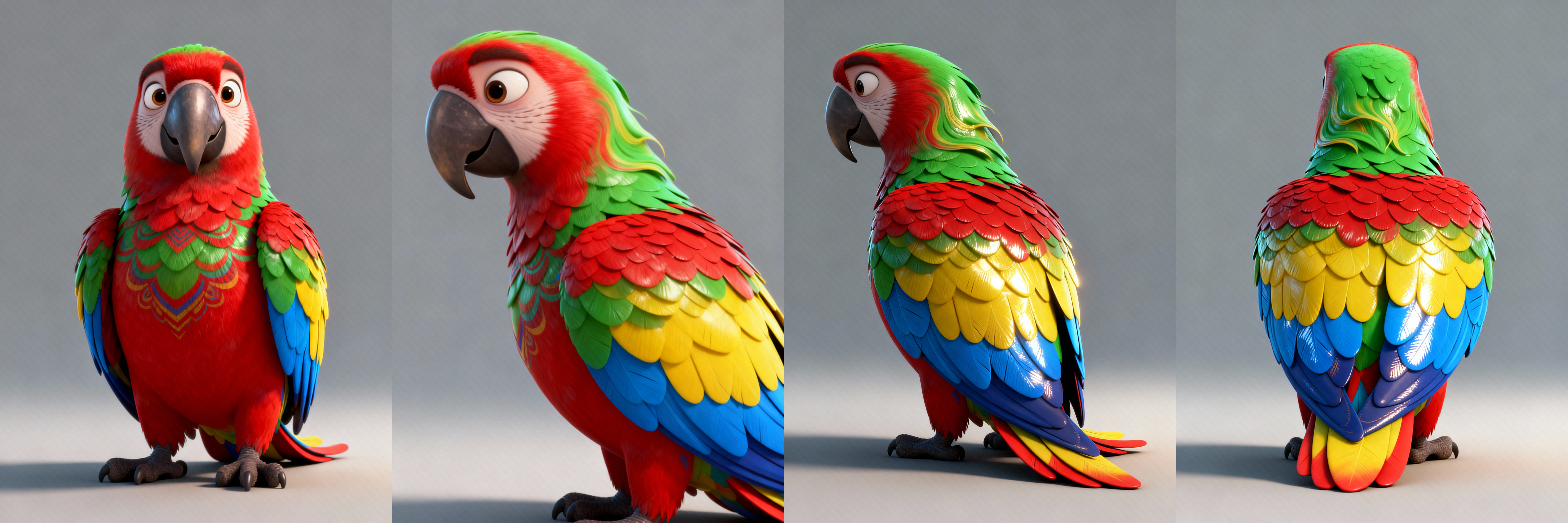}
\end{minipage}

\caption{Comparison of different methods for multi-image generation.}
\end{figure*}

\begin{figure*}[!hb]
\centering

\begin{tcolorbox}[colback=gray!10, colframe=black, boxrule=1pt, arc=2mm, 
                  left=5pt, right=5pt, top=3pt, bottom=3pt, width=0.95\textwidth]
    \centering
    \textbf{Prompt:} \small\textit{``Design product mockups featuring a retro, pixel art logo that reimagines our brand in an 8-bit style paired with a futuristic digital font. Apply the logo on 4 products: a portable gaming console, a vintage-style gaming t-shirt, a pixel art coffee mug, and a limited edition poster, using a monochromatic color scheme.''}
\end{tcolorbox}

\begin{minipage}[c]{0.08\textwidth}
    \centering
    \rotatebox{90}{\large\textbf{Ours}}
\end{minipage}%
\hspace{0.01\textwidth}%
\begin{minipage}[c]{0.89\textwidth}
    \centering
    \includegraphics[width=\textwidth]{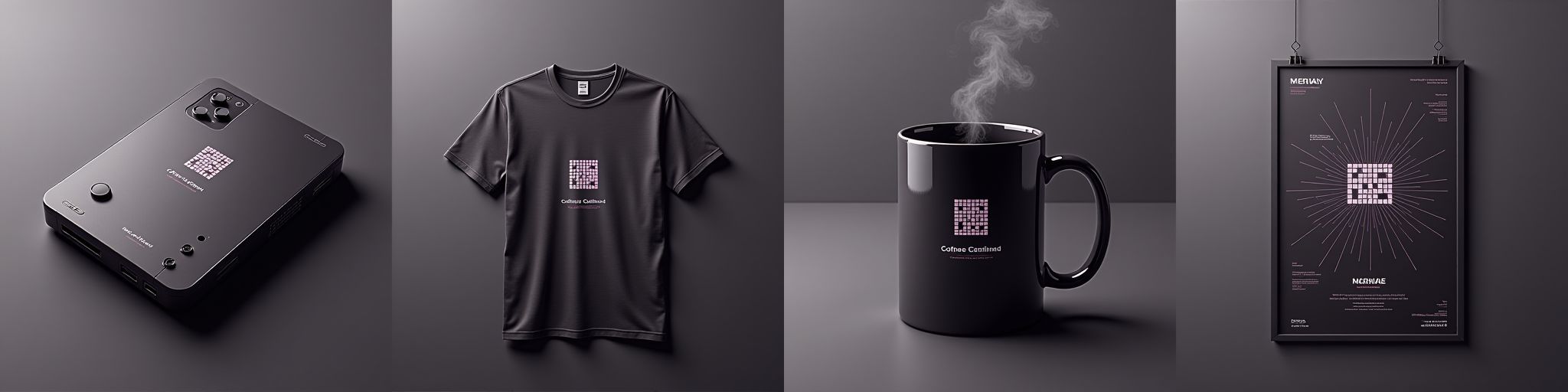}
\end{minipage}

\begin{minipage}[c]{0.08\textwidth}
    \centering
    \rotatebox{90}{\large\textbf{AutoT2IS}}
\end{minipage}%
\hspace{0.01\textwidth}%
\begin{minipage}[c]{0.89\textwidth}
    \centering
    \includegraphics[width=\textwidth]{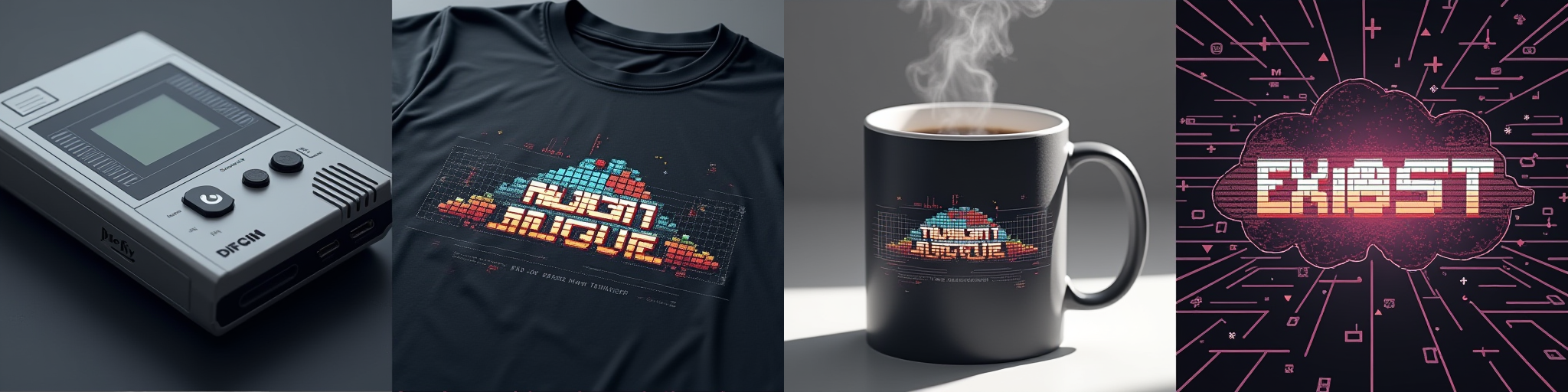}
\end{minipage}

\begin{minipage}[c]{0.08\textwidth}
    \centering
    \rotatebox{90}{\large\textbf{Seedream}}
\end{minipage}%
\hspace{0.01\textwidth}%
\begin{minipage}[c]{0.89\textwidth}
    \centering
    \includegraphics[width=\textwidth]{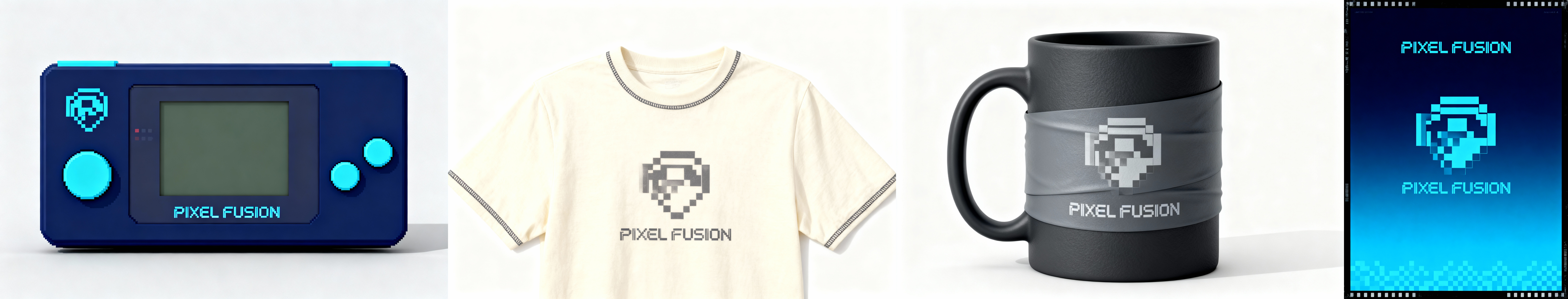}
\end{minipage}

\caption{Comparison of different methods for multi-image generation.}
\end{figure*}

\begin{figure*}[!hb]
\centering

\begin{tcolorbox}[colback=gray!10, colframe=black, boxrule=1pt, arc=2mm, 
                  left=5pt, right=5pt, top=3pt, bottom=3pt, width=0.95\textwidth]
    \centering
    \textbf{Prompt:} \small\textit{``This artwork represents the gradual creation of a traditional Chinese ink painting featuring pumpkins and vines. The progression follows these steps:...''}
\end{tcolorbox}

\begin{minipage}[c]{0.08\textwidth}
    \centering
    \rotatebox{90}{\large\textbf{Ours}}
\end{minipage}%
\hspace{0.01\textwidth}%
\begin{minipage}[c]{0.89\textwidth}
    \centering
    \includegraphics[width=\textwidth]{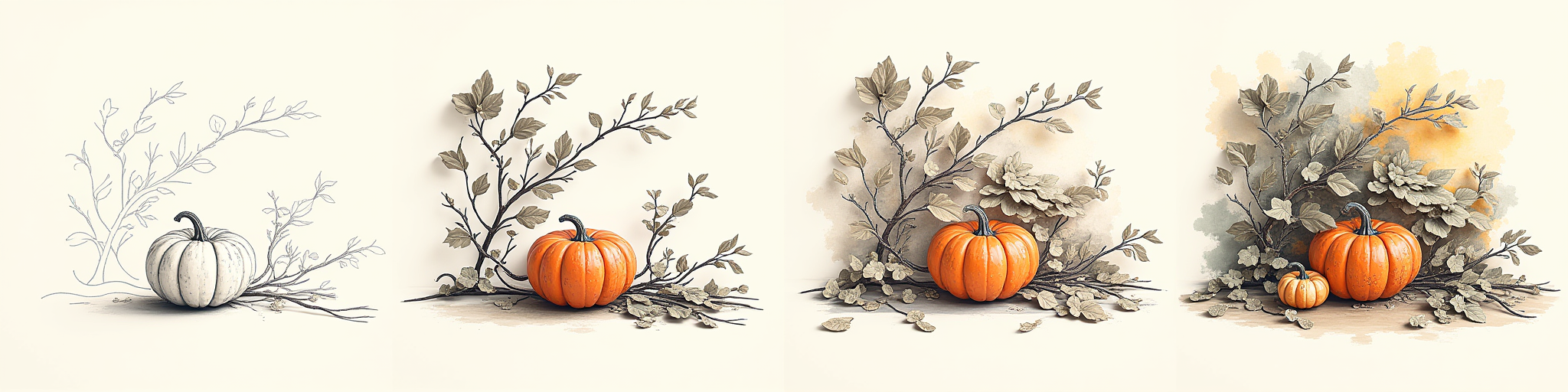}
\end{minipage}

\begin{minipage}[c]{0.08\textwidth}
    \centering
    \rotatebox{90}{\large\textbf{AutoT2IS}}
\end{minipage}%
\hspace{0.01\textwidth}%
\begin{minipage}[c]{0.89\textwidth}
    \centering
    \includegraphics[width=\textwidth]{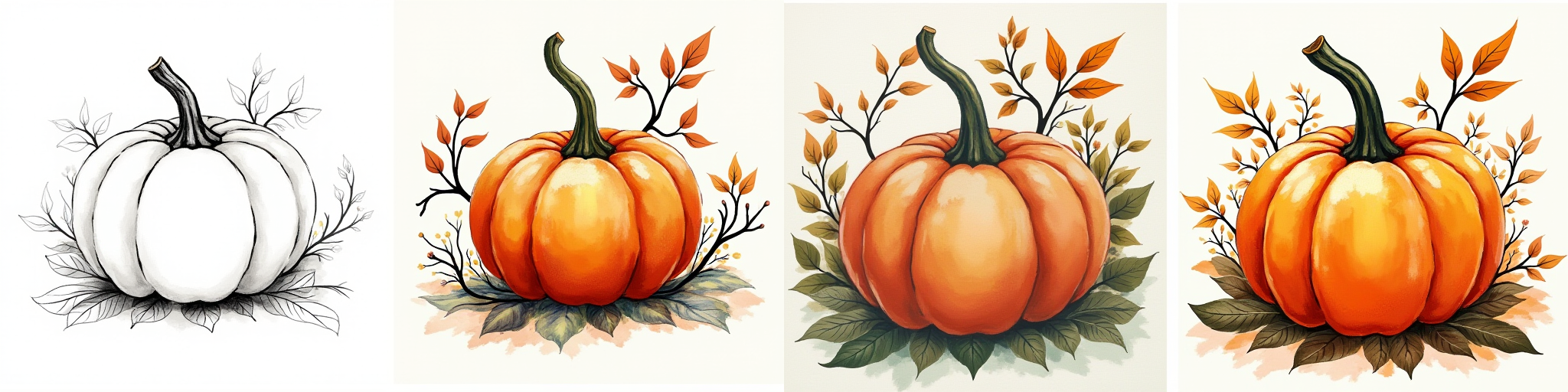}
\end{minipage}

\begin{minipage}[c]{0.08\textwidth}
    \centering
    \rotatebox{90}{\large\textbf{Seedream}}
\end{minipage}%
\hspace{0.01\textwidth}%
\begin{minipage}[c]{0.89\textwidth}
    \centering
    \includegraphics[width=\textwidth]{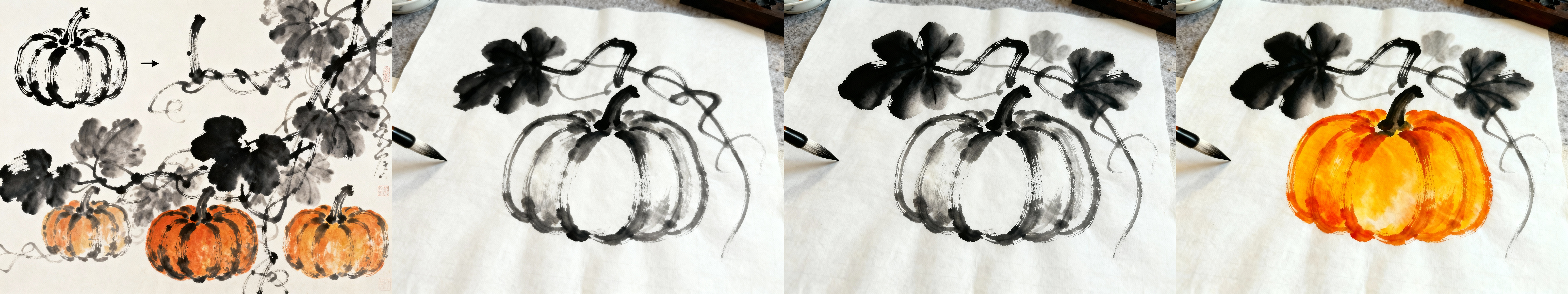}
\end{minipage}

\caption{Comparison of different methods for \textit{Text-to-ImageSet} generation.}
\end{figure*}

\clearpage
\setcounter{page}{1}
\maketitlesupplementary

\section{Implementation Details}\label{app:implementation_details}

\textbf{Infrastructure.}
All experiments are conducted on a server equipped with eight NVIDIA H100 GPUs, each with 80 GB of memory.

\textbf{PaCo-Reward Implementation Details.}
PaCo-Reward is fine-tuned using the LlamaFactory framework~\cite{zheng2024llamafactory} with a customized weighted cross-entropy loss defined in~\cref{eq:paco_loss}.
Both PaCo-Reward-7B-Fast and PaCo-Reward-7B are fine-tuned from Qwen2.5-VL-7B-Instruct~\cite{qwen2025qwen25technicalreport}, employing a LoRA rank of 32, $\alpha=64$,
a learning rate of $2\times10^{-4}$, and a batch size of 8.
Training on the PaCo-Dataset for two epochs requires approximately 18 GPU hours.
For PaCo-Reward-7B, the first-token weight $\alpha$ in~\cref{eq:paco_loss} is set to 0.1 to emphasize the importance of the initial token.
Since PaCo-Reward-7B-Fast is trained solely on binary labels (“Yes.”/“No.”), it does not apply token weighting.

\textbf{PaCo-RL Implementation Details.}
During PaCo-RL training, one GPU is dedicated to running a vLLM~\cite{kwon2023efficient} server for reward computation,
while the remaining seven GPUs are used for reinforcement-learning fine-tuning.
Unless otherwise specified, all experiments adopt a learning rate of $3\times10^{-4}$, a batch size of 1,
a group size of $G=16$, and 42 unique samples per epoch.
The clipping parameter $\varepsilon$ in~\cref{eq:grpo_objective} is set to $1\times10^{-4}$.
Following~\cite{liu2025flowgrpotrainingflowmatching}, we monitor KL loss throughout training but assign it zero weight ($\beta=0$) in~\cref{eq:grpo_objective} to achieve better performance.

\textit{Text-to-ImageSet.}
For the Text-to-ImageSet generation task,
we fine-tune FLUX.1-dev~\cite{blackforestlabs_flux1dev_2024} using LoRA with rank 64 and $\alpha=128$,
and apply a classifier-free guidance (CFG) scale of 3.5 during both training and inference.
The noise scale $\sigma_t$ in~\cref{eq:update_rule} is defined as $\sigma_t = a \sqrt{\tfrac{t}{1-t}}$ with $a=0.7$, following FlowGRPO~\cite{liu2025flowgrpotrainingflowmatching}.
We use 10 denoising steps and perform SDE sampling only at timestep $t=1$ (from ${0,1,\ldots,9}$),
leveraging MixGRPO~\cite{li2025mixgrpounlockingflowbasedgrpo} and FlowGRPO-Fast~\cite{liu2025flowgrpotrainingflowmatching} strategies for efficient training.
In our PaCo-RL setup, the training resolution is $512\times512$ while the evaluation resolution is $1024\times1024$.
We use $\delta=0.2$ in~\cref{eq:log_tamed_reward} to aggregate the Consistency Score (from PaCo-Reward-7B) and the Text-Image Alignment Score (from CLIP-T~\cite{radford2021learningtransferablevisualmodels}).

\textit{Image Editing.}
For Image Editing, we fine-tune both FLUX.1-Kontext-dev~\cite{batifol2025flux} and Qwen-Image-Edit~\cite{wu2025qwenimagetechnicalreport} using LoRA with rank 64 and $\alpha=128$.
The CFG scales are set to 2.5 and 4.0 for FLUX.1-Kontext-dev and Qwen-Image-Edit, respectively.
We apply the MixGRPO~\cite{li2025mixgrpounlockingflowbasedgrpo} strategy for efficient training,
setting the noise scale $a=0.9$ at timestep 1 for FLUX.1-Kontext-dev~\cite{batifol2025flux},
and $a=1.0$ for timesteps 1-4 for Qwen-Image-Edit~\cite{wu2025qwenimagetechnicalreport}.
The training and evaluation resolutions are $384\times384$ and $1024\times 1024$, respectively.
As the Image Editing task relies on a single reward signal, multi-reward aggregation and the log-tame strategy in~\cref{eq:log_tamed_reward} are not employed.
The prompt template for Image Editing reward computation is provided in~\cref{app:prompt_templates}.

\section{Prompt Templates}\label{app:prompt_templates}
We provide the prompt templates used for reward computation in both tasks below.

For the \textit{Text-to-ImageSet generation} task, we design two versions of the prompt template:
(1) one incorporating detailed consistency criteria derived from the original dataset for enhanced evaluation reliability, and
(2) another containing only the input prompt information, which is used for generalization evaluation.

For the \textit{Image Editing} task, we adopt a modified version of the prompt template from EditScore~\cite{luo2025editscoreunlockingonlinerl}.
This template consists of two components: \textit{Semantic Consistency (SC)} and \textit{Prompt Following (PF)}.

\begin{center}

\begin{tcolorbox}[
    title={\footnotesize Prompt Template for \textit{Text-to-ImageSet} (v1)},
    width=0.9\linewidth
]
\scriptsize

Do images meet the following criteria?

{\color{red!60}\{consistency\_criteria\}}

Please answer ``Yes'' or ``No'' first, then provide detailed reasons.

\end{tcolorbox}
\begin{tcolorbox}[
    title={\footnotesize Prompt Template for \textit{Text-to-ImageSet} (v2)},
    width=0.9\linewidth
]
\scriptsize
Given two subfigures generated based on the theme:

{\color{red!60}\{main\_prompt\}}

do the two images maintain consistency in terms of style, logic and identity?
Answer ``Yes'' or ``No'' first, and then provide detailed reasons.

\end{tcolorbox}
\begin{tcolorbox}[
    title={\footnotesize Prompt Template for \textit{Image Editing} (SC)},
    width=0.9\linewidth
]
\scriptsize
Compare the edited image (second) with the original image (first).
{
\color{red!60}
Instruction: \{prompt\}.
}
Except for the parts that are intentionally changed according to the instruction,
does the edited image remain consistent with the original in style, logic, and identity?
Answer 'Yes' or 'No' first, then provide detailed reasons.
\end{tcolorbox}
\begin{tcolorbox}[
    title={\footnotesize Prompt Template for \textit{Image Editing} (PF)},
    width=0.9\linewidth
]
\scriptsize
Compare the edited image (second) with the original image (first).

{\color{red!60}Instruction: \{prompt\}.}

Does the edited image accurately follow this instruction?
Answer 'Yes' or 'No' first, then provide detailed reasons.
\end{tcolorbox}

\end{center}
\clearpage
\section{Resolution-Decoupled Training Analysis}\label{app:resolution_analysis}

To validate the reliability of resolution-decoupled training, we conduct experiments generating image sets at three resolutions: $256\times256$ (0.25x), $512\times512$ (0.5x), and $1024\times1024$ (1x) using FLUX.1-dev, and analyze the Pearson correlation of evaluation metrics across resolutions.

\begin{figure}[htbp]
    \centering
    \begin{minipage}{0.48\linewidth}
        \includegraphics[width=\linewidth]{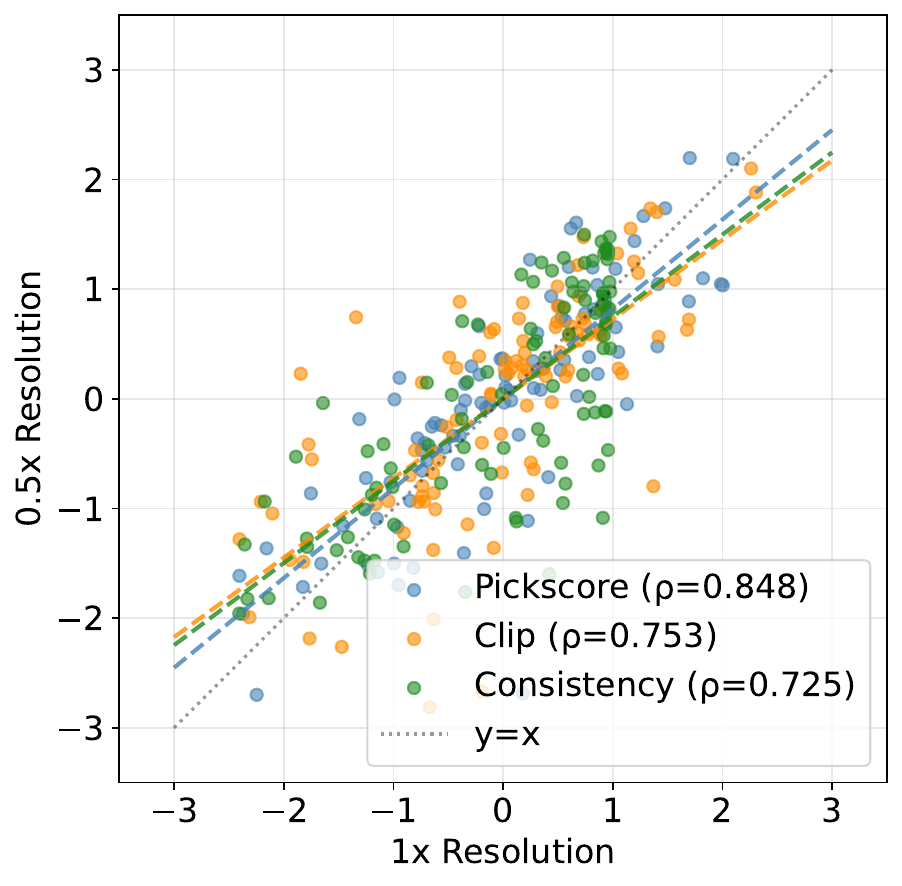}
        \caption*{(a) 1x vs.\ 0.5x}
    \end{minipage}
    \hfill
    \begin{minipage}{0.48\linewidth}
        \includegraphics[width=\linewidth]{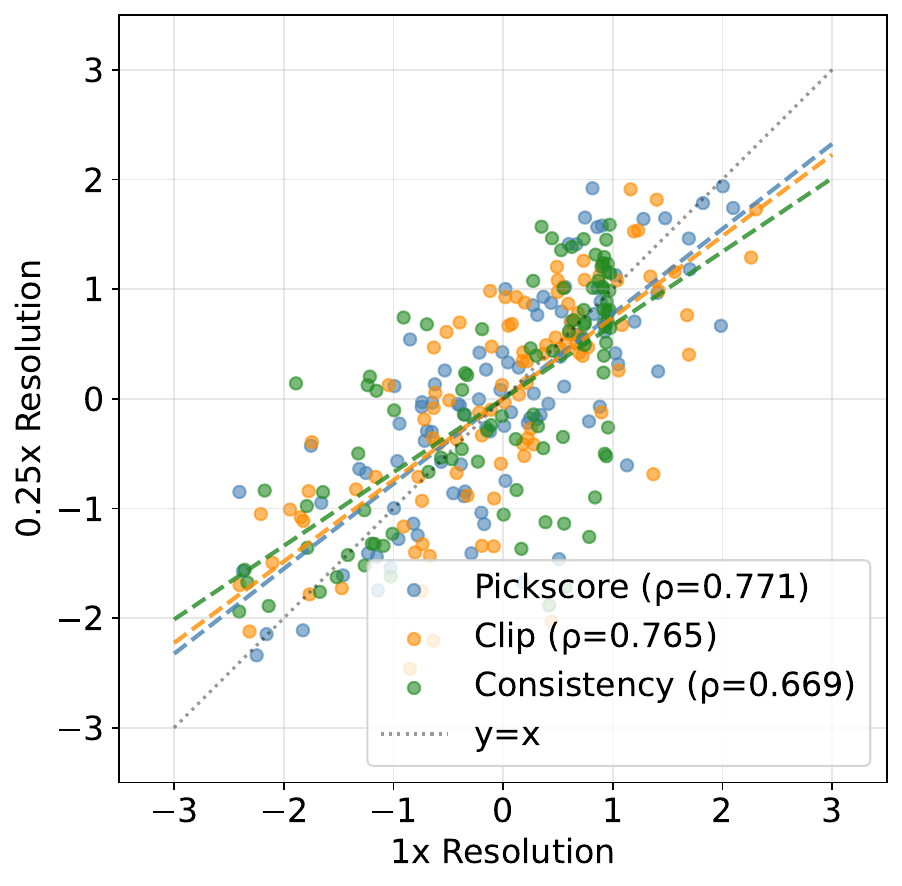}
        \caption*{(b) 1x vs.\ 0.25x}
    \end{minipage}
    \caption{Pearson correlation of evaluation metrics across different training-to-inference resolution ratios. Strong correlations at 0.5x confirm that reward signals remain reliable under moderate resolution reduction.}
    \label{fig:resolution_correlation}
\end{figure}

As shown in~\cref{fig:resolution_correlation}, all metrics exhibit strong positive correlations ($0.725$--$0.848$) when comparing 1x vs.\ 0.5x resolutions, indicating that the relative quality ordering of image sets is largely preserved and reward signals remain reliable under moderate resolution reduction. However, correlations weaken substantially when comparing 1x vs.\ 0.25x, especially for Aesthetics and Consistency, suggesting that extremely low resolutions lose detailed visual information crucial for these fine-grained aspects.

We further validated these findings on additional reward models (Text-Rendering, GenEval) and generators (Qwen-Image, FLUX2), confirming that 0.5x reduction consistently preserves reward reliability while 0.25x fails, particularly for fine-grained metrics. Regarding higher-resolution inference, no systematic artifacts were observed when generating at full resolution after 0.5x training, as the reward improvements transfer effectively across resolutions. This supports the use of resolution-decoupled training as a practical strategy for reducing computational cost without sacrificing final generation quality.

\section{Ablation on PaCo-GRPO Components}\label{app:paco_grpo_ablation}

We quantify the individual contributions of two key components in PaCo-GRPO: resolution-decoupled training (Res-Dec.) and log-tamed reward aggregation (Log-Agg.). Results are reported in~\cref{tab:paco_grpo_ablation}.

\begin{table}[htbp]
    \centering
    \caption{Ablation study on PaCo-GRPO components. Removing resolution-decoupled training leads to suboptimal performance even with doubled training time, while removing log-tamed aggregation causes reward collapse.}
    \label{tab:paco_grpo_ablation}
    \setlength{\tabcolsep}{10pt}
    \resizebox{\columnwidth}{!}{
    \begin{tabular}{lcccc}
        \toprule
        Method & Aes.$\uparrow$ & P.F.$\uparrow$ & V.C.$\uparrow$ & Time$\downarrow$ \\
        \midrule
        Full PaCo-GRPO          & \textbf{0.555} & \textbf{0.728} & 0.493          & \textbf{6.0h} \\
        W/O Res-Dec.            & 0.542          & 0.698          & 0.452          & 12.0h         \\
        W/O Log-Agg.            & 0.471          & 0.616          & \textbf{0.557} & 6.0h          \\
        \bottomrule
    \end{tabular}}
\end{table}

\paragraph{Resolution-Decoupled Training.}
Removing resolution-decoupled training results in suboptimal performance even with doubled training time (12h vs.\ 6h), as the model fails to converge fully at higher resolution. This demonstrates that training at a reduced resolution (0.5x) not only halves the computational cost but also leads to better optimization dynamics by enabling more effective exploration within the same time budget.

\paragraph{Log-Tamed Reward Aggregation.}
Without log-tamed aggregation, the visual consistency reward dominates the training signal, causing the model to collapse into generating near-identical, low-quality image sets. While the Visual Consistency (V.C.) score increases to 0.557, both Aesthetics and Prompt Following degrade significantly, confirming that the log-tamed strategy in~\cref{eq:log_tamed_reward} is essential for balancing multiple reward objectives and preventing reward hacking.

\section{PaCo-Dataset Details}\label{app:paco_dataset_details}

To ensure the quality and consistency of PaCo-Dataset annotations,
all annotators followed detailed guidelines accompanied by illustrative examples to establish a shared understanding of the evaluation criteria across different consistency dimensions (\ie, style, logic, and identity).
Each image pair was independently evaluated by multiple annotators,
and the final annotations were obtained via majority voting.
Ties were discarded to ensure clear binary preferences in the dataset.

The main categories, subcategories, and their corresponding consistency dimensions are summarized in~\cref{tab:paco_dataset_details}.

\begin{table*}[!hb]
\centering
\caption{Main categories, subcategories, and their corresponding consistency dimensions.}
\label{tab:paco_dataset_details}
\begin{tabularx}{\textwidth}{llX}
\toprule
\textbf{Main Category} & \textbf{Subcategory} & \textbf{Consistency Dimensions} \\
\midrule
\multirow{5}{*}{Design Style Generation} 
& Home Decoration & Style \\
& IP Product & Style, Identity \\
& Font Design & Style \\
& Poster Design & Style, Logic \\
& Creative Style & Style \\
\midrule
\multirow{5}{*}{Story Generation}
& Children Book & Logic, Identity, Style \\
& Hist. Narrative & Logic, Identity \\
& Movie Shot & Logic, Identity, Style \\
& Comic Story & Logic, Identity, Style \\
& News Illustration & Logic, Style \\
\midrule
\multirow{6}{*}{progression Generation}
& Evolution Illustration & Logic \\
& Draw progression & Logic, Style \\
& Growth progression & Logic \\
& Arch. Building & Logic \\
& Cooking progression & Logic \\
& Physical Law & Logic \\
\midrule
\multirow{6}{*}{Instruction Generation}
& Historical Panel & Logic, Style \\
& Activity Arrange & Logic \\
& Evolution Illustration & Logic \\
& Education Illustration & Logic, Style \\
& Travel Guide & Logic, Style, Identity \\
& Product Instruction & Logic, Style \\
\midrule
\multirow{5}{*}{Character Generation}
& Multi-view & Identity, Style \\
& Multi-pose & Identity \\
& Portrait Design & Identity, Style \\
& Multi-Expression & Identity \\
& Multi-Scenario & Identity, Logic \\
\midrule
\multirow{6}{*}{Editing}
& Inpainting and replacement & Identity \\
& Element manipulation & Identity, Style \\
& Background modification & Identity, Style, Logic \\
& Attribute and effect manipulation & Style \\
& Image editing and manipulation & Identity, Style, Logic \\
\bottomrule
\end{tabularx}
\end{table*}

\end{document}